%% file: main.tex
\documentclass[10pt,twocolumn,letterpaper]{article}

\usepackage[pagenumbers]{cvpr}

\usepackage{graphicx}
\usepackage{amsmath}
\usepackage{amssymb}
\usepackage{booktabs}
\usepackage{marvosym}
\usepackage{xcolor}
\usepackage{algorithm}
\usepackage{algpseudocode}
\usepackage{multirow}
\usepackage{color}
\usepackage[pagebackref,breaklinks,colorlinks]{hyperref}
\usepackage[capitalize]{cleveref}

\crefname{section}{Sec.}{Secs.}
\Crefname{section}{Section}{Sections}
\Crefname{table}{Table}{Tables}
\crefname{table}{Tab.}{Tabs.}

\let\titleold\title
\renewcommand{\title}[1]{\titleold{#1}\newcommand{\thetitle}{#1}}
\def\maketitlesupplementary
   {
   \newpage
       \twocolumn[
        \centering
        \Large
        \textbf{\thetitle}\\
        \vspace{0.5em}Supplementary Material \\
        \vspace{1.0em}
       ] 
   }
\title{MovieChat: From Dense Token to Sparse Memory for Long Video Understanding}

\author{
Enxin Song\textsuperscript{1,\thanks{Equal contribution, \textsuperscript{$\dagger$}~Project lead, \textsuperscript{$\ddagger$}~Data collection, \textsuperscript{\Letter}~Corresponding Author.}} \quad
Wenhao Chai\textsuperscript{2,$*$,$\dagger$} \quad
Guanhong Wang\textsuperscript{1,$*$} \quad
Yucheng Zhang\textsuperscript{1,$\ddagger$} \quad \\
Haoyang Zhou\textsuperscript{1,$\ddagger$} \quad
Feiyang Wu\textsuperscript{1,$\ddagger$} \quad
Haozhe Chi\textsuperscript{1} \quad
Xun Guo\textsuperscript{3} \\
Tian Ye\textsuperscript{4} \quad
Yanting Zhang\textsuperscript{5} \quad
Yan Lu\textsuperscript{3} \quad
Jenq-Neng Hwang\textsuperscript{2} \quad
Gaoang Wang\textsuperscript{1,\Letter}\\
[2mm]
\textsuperscript{1}~Zhejiang University \quad \textsuperscript{2}~University of Washington \quad \textsuperscript{3}~Microsoft Research Asia\\
\textsuperscript{4}~Hong Kong University of Science and Technology (GZ) \quad
\textsuperscript{5}~Donghua University
\\
[2mm]
\normalsize{
\url{https://rese1f.github.io/MovieChat}
}
}

\begin{document}

\maketitle
\input{tex_arxiv/0_abs}
\input{tex_arxiv/1_intro}
\input{tex_arxiv/2_survey}
\input{tex_arxiv/3_method}
\input{tex_arxiv/4_dataset}

\input{tex_arxiv/5_exp}
\input{tex_arxiv/6_conclusion}
\newpage
{\small
\bibliographystyle{ieee_fullname}
\bibliography{main}
}
\maketitlesupplementary
\input{tex_arxiv/7_appendix}
\end{document}

%% file: tex_arxiv/0_abs.tex
\begin{abstract}
Recently, integrating video foundation models and large language models to build a video understanding system can overcome the limitations of specific pre-defined vision tasks. Yet, existing systems can only handle videos with very few frames. For long videos, the computation complexity, memory cost, and long-term temporal connection impose additional challenges. Taking advantage of the Atkinson-Shiffrin memory model,  with tokens in Transformers being employed as the carriers of memory in combination with our specially designed memory mechanism,  we propose the MovieChat to overcome these challenges. MovieChat achieves state-of-the-art performance in long video understanding, along with the released MovieChat-1K benchmark with 1K long video and 14K manual annotations for validation of the effectiveness of our method.
\end{abstract}

%% file: tex_arxiv/1_intro.tex
\section{Introduction}

Recent advances in Large Language Models (LLMs)~\cite{gpt4,brown2020language,touvron2023llama,chiang2023vicuna,taori2023stanford} acheive great success in Natural Language Processing (NLP) . It is a natural progression to introduce multi-modality~\cite{chai2022deep} into LLMs and turn it into Multi-modal Large Language Models (MLLMs), which is able to conduct multimodal rationalization and understanding. MLLMs have shown incredible emergent capabilities in various multimodal tasks such as perception (\eg, existence, count, position, OCR)~\cite{wang2023visionllm, maaz2023video, alayrac2022flamingo,zhu2023minigpt,li2023otter,li2023blip}, commonsense reasoning~\cite{gao2023llama,zhu2023minigpt,li2023otter,li2022blip,li2023blip,gong2023multimodal,su2023pandagpt, maaz2023video}, and code reasoning~\cite{fu2023mme,lyu2023macaw,ye2023mplug,dai2023instructblip,liu2023visual,gao2023llama}, resulting in a potential path to Artificial General Intelligence (AGI). Compared to LLMs and other task-specific models, MLLMs  provide a more human-like interpretation of the scenarios, a user-friendly interface for interaction, and a broader range of capabilities.

\input{fig/intro}

Existing vision-centric MLLMs follow the paradigm that utilizing pre-trained LLMs and visual encoder with additional learnable modules (Q-former~\cite{dai2023instructblip,li2022blip,li2023blip,zhang2023video} or simple projection layer~\cite{driess2023palm,maaz2023video,liu2023visual,su2023pandagpt}). In video field, some previous works~\cite{zhang2023video,maaz2023video} follow this paradigm to build video MLLMs, while works in the other paradigm~\cite{wang2023chatvideo,li2023videochat} combine existing visual perception tools (\eg, tracking and classification) and LLMs through Application Programming Interface (API) to build a system without training. Yet, previously, there is no exploration of a model or system based on long videos (over one minute), and there is also a lack of a standardized benchmark to evaluate the capabilities of these systems.

In this paper, we present MovieChat, a novel framework that integrates vision models and LLMs to conduct long video understanding tasks. We claim that the computation complexity, memory cost, and long-term temporal connection are the main challenges for long video understanding. Atkinson-Shiffrin memory model~\cite{atkinson1968chapter} proposes that short-term memory functions as a buffer of long-term memory, serving as a processor for the encoding of information into long-term memory. Inspired by this, we propose a memory mechanism to deal with long video understanding tasks, which includes a rapidly updated short-term memory and a compact thus sustained long-term memory. We use a sliding window approach to extract video features and represent them in token form, which are then sequentially fed into the short-term memory frame by frame. The short-term memory has a fixed length, and when it reaches its set limit, the earliest tokens are popped and consolidated into the long-term memory. After passing through a projection layer, the video representation is inputted into a large language model for interaction with the user. As shown in Fig.~\ref{fig:intro}, our proposed MovieChat mechanism outperforms other existing methods in terms of Video Random Access Memory (VRAM) cost. We also release a new benchmark, MovieChat-1K, with 1K long videos and 13K manual question-answering pairs for validation of the effectiveness of our proposed MovieChat. 

The contributions of this work are summarized as:

\begin{itemize}
    \item We present MovieChat, a novel framework that integrates vision models and LLMs, which is the first to support long video ($\textgreater 10$K frames) understanding tasks.
    \item We propose an effective memory management mechanism to reduce the computation complexity and memory cost, while enhancing the long-term connection.
    \item We release the first long video understanding benchmark, MovieChat-1K, with manual annotations and conduct extensive quantitative evaluation and case studies to evaluate the comparable performance of both understanding capability and inference cost.
\end{itemize}

%% file: fig/intro.tex
\begin{figure}[t]
    \centering
    \includegraphics[width=1\linewidth]{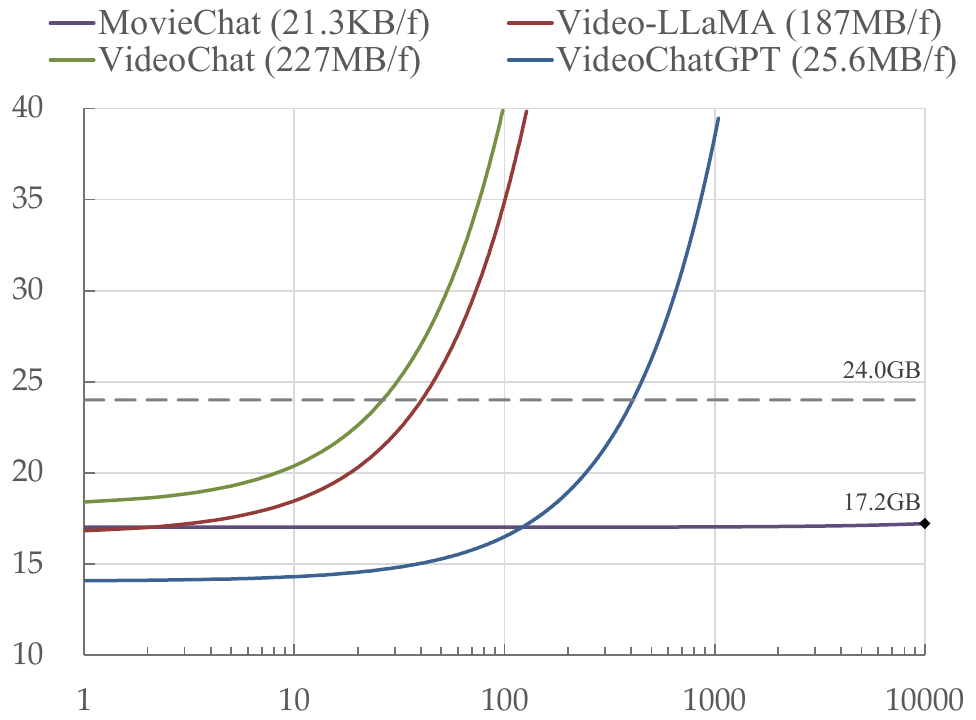}
    \caption{VRAM cost under gigabyte (GB) (y-axis) v.s. frame number (x-axis) comparison. We test the visual-only inference of all methods at a resolution of $224 \times 224$ without frame sampling. While the previous method can only support around $100$ frames of inference, MovieChat can handle videos with $\textgreater 10$K frames on a 24GB graphics card. MovieChat has a $10000 \times$ advantage over other methods in terms of the average increase in VRAM cost per frame ($21.3$KB to $\sim200$MB per frame).}
    \label{fig:intro}
\end{figure}

%% file: tex_arxiv/2_survey.tex
\section{Related Works}

\subsection{Multi-modal Large Language Models}
LLMs~\cite{gpt4,brown2020language,touvron2023llama,touvron2023llama2,chiang2023vicuna,taori2023stanford} have achieved great success in natural language processing (NLP) tasks recently. Many works try to build MLLMs~\cite{alayrac2022flamingo,zhu2023minigpt,li2023otter,li2023blip,gong2023multimodal,ye2023mplug,dai2023instructblip,wang2023visionllm,maaz2023video,gao2023llama} by combining models of other modalities.
Flamingo~\cite{alayrac2022flamingo} bridges powerful pre-trained vision-only and language-only models and achieves state-of-the-art performance with few-shot learning.
BLIP-2~\cite{li2023blip} proposes a generic and efficient pre-training strategy that bootstraps vision-language pre-training from an off-the-shelf frozen pre-trained image encoders and a frozen large language model.
MiniGPT-4~\cite{zhu2023minigpt} also aligns a frozen visual encoder with a frozen LLM, Vicuna~\cite{chiang2023vicuna}, using just one projection layer to realize the system.
Otter~\cite{li2023otter} showcases improved instruction-following ability and in-context learning.
In video field, ChatVideo~\cite{wang2023chatvideo} treats tracklets as the basic video unit and allows users' interacting with the LLMs.
VideoChat~\cite{li2023videochat} integrates video foundation models and LLMs via a learnable neural interface, excelling in spatiotemporal reasoning, event localization, and causal relationship inference. 
Video-LLaMA~\cite{zhang2023video} further leverages pre-trained models ImageBind~\cite{girdhar2023imagebind} and LLaMA~\cite{touvron2023llama}, bootstraping cross-modal training in videos following BLIP-2.
Yet, these methods fail to handle long video understanding because of high computation complexity, large memory cost, and weak long-term temporal connection. Therefore, our main effort is to introduce an effective memory mechanism to overcome these challenges.

\input{fig/overview}

\subsection{Long Video Understanding}

Understanding long videos is a challenging task in computer vision. Prior arts use 3D CNN for long-term feature bank~\cite{wu2019long}, object/human-centric motion~\cite{wu2021towards,rohrbach2017generating}, or other forms~\cite{wu2022memvit,sener2020temporal} as video representations. MIST~\cite{gao2023mist} decomposes dense self-attention into a cascade segment and
region selection module to increase the computation efficiency for understanding minutes of long videos. Building long-form video understanding datasets is challenging and rarely explored.  \cite{shou2021generic} captures large scale data from Kinetics-400~\cite{carreira2017quo}, but only for generic event boundary detection tasks. \cite{soldan2022mad} creates a language grounding benchmark from audio descriptions of movies, but it lacks long-term understanding evaluation. \cite{tapaswi2016movieqa} successfully builds a benchmark contains multiple sources of information (\eg, video clips, plots, and DVS) for question-answering tasks in the movie field. There are also several datasets of video-caption/description pairs among various domains, such as cooking (\eg, MPII Cooking~\cite{rohrbach2012database,rohrbach2012script,rohrbach2016recognizing} and TACoS~\cite{regneri2013grounding,rohrbach2014coherent}), instruction (\eg, HowTo100M~\cite{miech2019howto100m} and HiREST~\cite{zala2023hierarchical}), Ego~\cite{mangalam2023egoschema}, and movie (\eg, MovieQA~\cite{tapaswi2016movieqa} and MovieNet~\cite{huang2020movienet}) from different sources such as YouTube~\cite{chen2011collecting,zeng2016title,miech2019howto100m}, Twitter~\cite{awad2017trecvid,awad2018trecvid,awad2020trecvid,awad2021trecvid}, and Internet~\cite{bain2021frozen}. 
Yet, those datasets lack diverse and fine-grained dense captioning for long videos.

\subsection{Memory Models in Vision Tasks}

There are some prior works exploring memory models~\cite{squire2015memory} in various vision tasks in videos, such as video object segmentation~(VOS)~\cite{cheng2022xmem,hu2021learning,seong2020kernelized,seong2021hierarchical}, multi-object tracking~(MOT)~\cite{cai2022memot,hao2022umotma,xin2022multi,allen2006multiple}, visual object tracking~(VOT)~\cite{zhou2023memory,yang2018learning,liu2017mavot,ma2018adaptive}, and action understanding~\cite{wang2023memory}.
MeMOT~\cite{cai2022memot} builds a large spatiotemporal memory that stores the past observations of the
tracked objects. 
XMem~\cite{cheng2022xmem} develops an architecture that incorporates multiple independent yet deeply-connected feature memory storage to handle long videos with thousands of frames.
We learn from the experience of those prior arts and further adopt an effective memory mechanism in combination with LLMs. 

Our method focuses on reducing the redundancy of visual tokens in the video and building a memory mechanism to pass the information among a large temporal range.

%% file: fig/overview.tex
\begin{figure*}[t]
    \centering
    \includegraphics[width=1\linewidth]{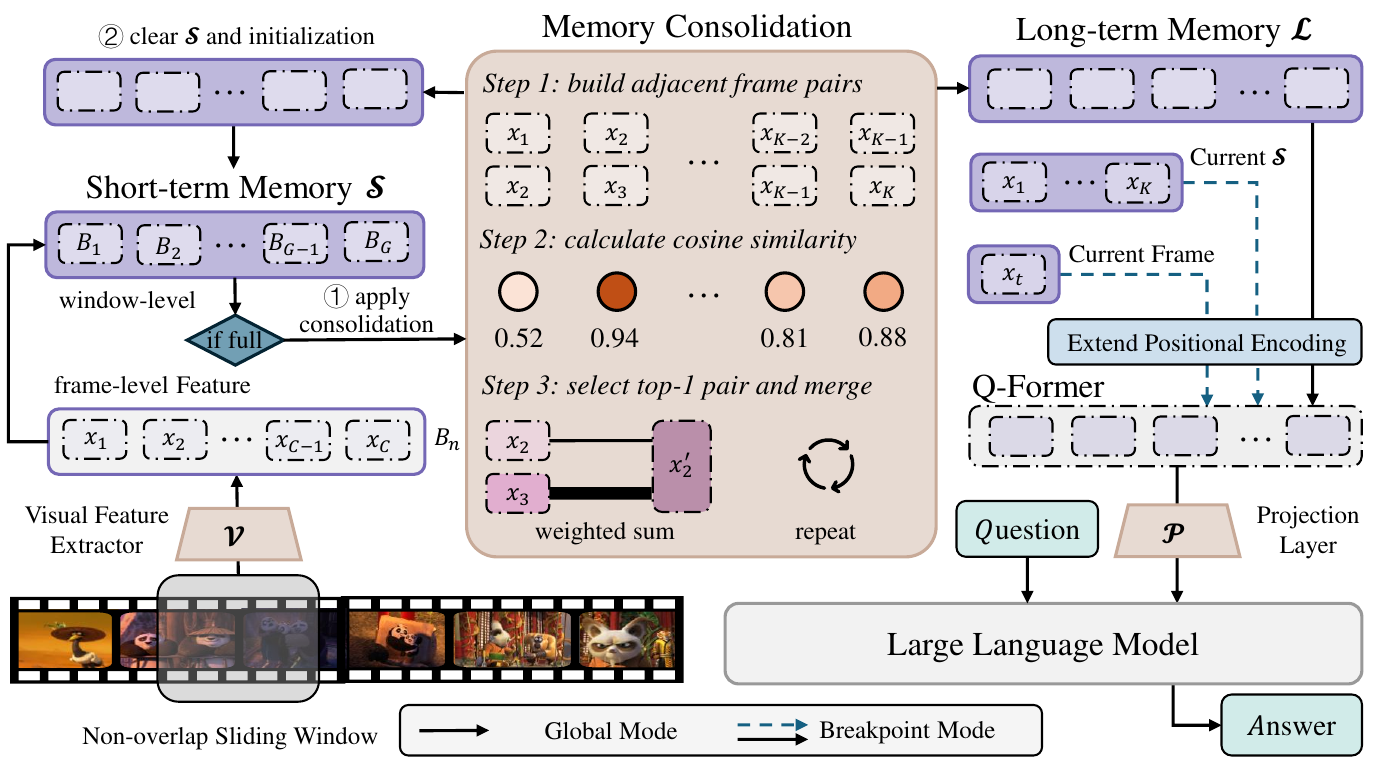}
    \caption{\textbf{Illustration of MovieChat.} MovieChat extracts video features with a sliding window and represents them in token form, which are then sequentially fed into the short-term memory frame by frame. When the fixed-length short-term memory reaches its preset limit, the earliest tokens are popped and consolidated into the long-term memory. MovieChat incorporates two distinct inference modes: the global mode, which exclusively utilizes the long-term memory, and the breakpoint mode, which additionally incorporates the current short-term memory as part of the video representation. The breakpoint mode allows for understanding the video at a specific moment in time. After passing through a projection layer, the video representation is inputted into a large language model for interaction with the user.}
    \label{fig:overview}
\end{figure*}


%% file: tex_arxiv/3_method.tex
\section{MovieChat}

\subsection{Overview}
Our proposed method, MovieChat, comprises several key components, including the frame-wise visual feature extractor, the short-term and long-term memory modules, the video projection layer, and the Large Language Model (LLM), as illustrated in Fig.~\ref{fig:overview}. MovieChat is designed for ultra-long videos ($\textgreater 10$K frames) understanding through interactive dialogue with the user. To address the impractical storage demands of concurrently storing a vast number of frames in both GPU memory and RAM, we employ a sliding window approach to efficiently process the video. The short-term memory module embeds dense tokens with sliding window and the long-term memory module periodically updates. MovieChat supports two inference modes: Breakpoint mode is used to understand a specific moment in the video, providing insights and answers based on that particular frame or scene; Global mode, on the other hand, is employed to comprehend the entire video as a whole, enabling a comprehensive understanding of the overall content and context.

\subsection{Visual Feature Extraction}
For visual feature extraction, instead of utilizing video-based foundational models such as ViViT~\cite{arnab2021vivit} or Video-Swin~\cite{liu2022video}, we simply use an image-based model to get frame-wise feature in the form of tokens. To be specific, we utilize pre-trained models as our visual feature extractor, including the ViT-G/14 from EVA-CLIP~\cite{fang2022eva} and the Q-former from BLIP-2~\cite{li2023blip2}. This is mainly because 1) there is few video foundation model that makes good alignment with text, and 2) our proposed memory mechanism can effectively capture temporal features. Given a raw video, the visual input $\mathbf{v} \in \mathbb{Z}^{T \times 3 \times H \times W}$ is a sequence of $T$ RGB frames of size $H \times W$ sampled from the video. The visual features are extracted in a sliding window manner, which could be formulated as
\begin{equation}
    B_{n} = \{ \mathbf{x}_{i}= \mathcal{V}(\mathbf{v}_{i}) \mid \forall i = 1,...,C\}, n = 1,...,\lceil \frac{T}{C} \rceil,
\end{equation}
where $B_{n}$ is the $n$-th video clip feature within the sliding window spanning $C$ frames. $\mathcal{V}(\cdot)$ is the visual feature extractor, taking as input a single frame {$\mathbf{v}_{i}\in \mathbb{Z}^{3 \times H \times W}$}. {$\mathbf{x}_{i}$} $\in \mathbb{R}^{N \times D}$ denotes $N$ extracted visual tokens with respect to each frame, and $D$ is the feature dimension of each token.

\subsection{Short-term Memory}
Short-term memory stores the frame tokens in a temporary fixed-length buffer. The previously extracted visual features by sliding window $G$ times without further processing are used to construct short-term memory, which can be formulated by:
\begin{equation}
    \mathcal{S} = \bigcup_{n}{B}_{n} = \{ \mathbf{{x}}_{i} \mid \forall i = 1, ..., K\}, n=1,..,G,
\end{equation}
where $\mathcal{S}$ is short-term memory, and $K$ is equal to $ C \times G$. Note that we set short-term memory to contain a fixed length of $K$ frames since the role of short-term memory is to assist in video understanding based on previous short-term contextual information. 

The update strategy for short-term memory is based on the First-in-First-out~(FIFO) queue. As a new batch of visual tokens enters, when the short-term memory reaches its capacity, we pop the currently stored frames to the memory consolidation module and clear the short-term memory. The output video feature obtained from the consolidation module augments the long-term memory; on the other hand, it reinitializes the short-term memory with this feature. The initialization aims at communicating the information between different sliding windows, thereby achieving more efficient compression.

\subsection{Long-term Memory}

Long-term memory can effectively avoid the problem of catastrophic knowledge forgetting, which is crucial for handling long video understanding tasks. The features stored in short-term memory are dense tokens, but due to the limitations of GPU memory and computation cost, storing all the tokens dropped from short-term memory into long-term memory buffer in sequence is infeasible. Besides, we observe significant temporal redundancy in videos, where activities span multiple frames with minimal visual changes. To this end, we propose a method to merge adjacent similar frames to simplify video feature representation and accelerate video encoding. This method transforms the dense tokens to the sparse memories, which are stored in long-term memory. 

\input{alg/longterm}
\input{fig/benchmark}

\input{fig/wordcloud}

To be specific, as shown in Algorithm~\ref{alg:longterm}, we conduct memory consolidation by merging the most similar tokens in the adjacent frames following ToMe~\cite{bolya2022token} periodically. We find that the token embedding in Transformer already summarize the information of each frame for using in calculating the average cosine similarity $s$ of $N$ embedded tokens:
\begin{equation}
    s = \frac{1}{N} \sum_{j=1}^{N} \left [ \cos(\mathbf{x}_{i}^{j}, \mathbf{x}_{i+1}^{j}) \right ].
\end{equation}

Our goal is to keep $R_{L}$ frames after every merge operation, which also embeds rich information stored in the long-term memory. $R_{L}$ is the hyper-parameter to control the trade-offs between performance and efficiency. Therefore, we greedily merge each set of adjacent frames with the highest similarity via weighted averaging. The merge operation is iteratively conducted until the token count reaches the predefined value set $R_{L}$ for each consolidation operation, resulting in the output video feature $\mathbf{v'} \in \mathbb{Z}^{R_{L}\times 3 \times H \times W}$ (operational details in appendix). The above algorithm is parameter-free, and can be easily plugged into a frame-based video encoder. Although the frame similarity calculation brings additional computing overhead, it is negligible compared to the efficiency gained by reducing stored frames.

\paragraph{Extend positional encoding.} For long-term memory, the number of tokens exceeds the maximum length of the positional encoding from the pre-trained model. Thus, our model utilizes the positional encoding mechanism following BERT~\cite{kenton2019bert}, which results in a portion exceeding the length threshold $n$ without available positional encoding. In order to handle long enough long memory, we adopt the hierarchically decomposed positional encoding method proposed by Su~\etal~\cite{pos}, which allows to extend the absolute positional encoding of length from $n$ to $n^2$.

\subsection{Inference}

Previous methods always use the representation of the whole video to conduct understanding and question-answering, which may fail in localizing specific moment especially in long videos. To this end, we propose two inference modes, global and breakpoint, for long video understanding task as follows.

\paragraph{Global mode.}

Global mode is defined as the understanding and question-answering for the whole video. In this case, we only use long-term memory $\mathcal{L}$ as the video representation $\mathbf{V}$.

\paragraph{Breakpoint mode.}

Breakpoint mode is distinctly defined as understanding specific moments in a video. Since events inherently possess continuity, we need to consider not only the information directly related to the moments stored in short-term memory $\mathcal{S}$ but also the information indirectly related stored in long-term memory $\mathcal{L}$. Based on this, we hypothesize that when querying the movie at a specific moment $t$, the video representation $\mathbf{V}$ should be the aggregation of $\mathcal{L}$, $\mathcal{S}$, and the current video frame feature $\mathbf{x}_{t}$. We find that simply concatenating these items yields excellent performance and leave further exploration of additional aggregation choices for future work.

Subsequently, the video representation $\mathbf{V}$ goes through a Q-former and a linear projection layer before being fed into the LLM $\mathcal{O}$, which can be formulated as:
\begin{equation}
    \mathbf{A} = \mathcal{O}(\mathbf{Q}, \mathcal{P} (\mathbf{V})),
\end{equation}
where $\mathcal{P}$ is the projection from visual space to text space. $\mathbf{A}$ represents the answer or instruction, and $\mathbf{Q}$ is employed to denote the question, respectively.

%% file: alg/longterm.tex
\definecolor{global}{RGB}{21,96,130}
\definecolor{breakpoint}{RGB}{51,0,111}

\begin{algorithm}[t]
\caption{Memory consolidation}\label{alg:longterm}
\begin{algorithmic}[1]
\Require $\mathcal{S}$ \textcolor{global}{\Comment{short-term memory}}
\While{$len(\mathcal{S}) \textgreater R_{L}$} \textcolor{global}{\Comment{iterative merge}}
\For{$\mathbf{x}_{i}$ in $\mathcal{S}$}
\State $s \gets sim(\mathbf{x}_{i}, \mathbf{x}_{i+1})$ \textcolor{global}{\Comment{tokens similarity}}
\EndFor
\State $m \gets max(s)$ \textcolor{global}{\Comment{the maximum value index}}
\State $\mathbf{x}_{m} \gets merge (\mathbf{x}_{m},\mathbf{x}_{m+1})$ \textcolor{global}{\Comment{merge}}
\State \textbf{del} $\mathbf{x}_{m+1}$
\EndWhile
\end{algorithmic}
\end{algorithm}

%% file: fig/benchmark.tex
\begin{figure*}[t]
  \centering

  \begin{subfigure}{0.33\textwidth}
    \centering
    \includegraphics[width=0.95\textwidth]{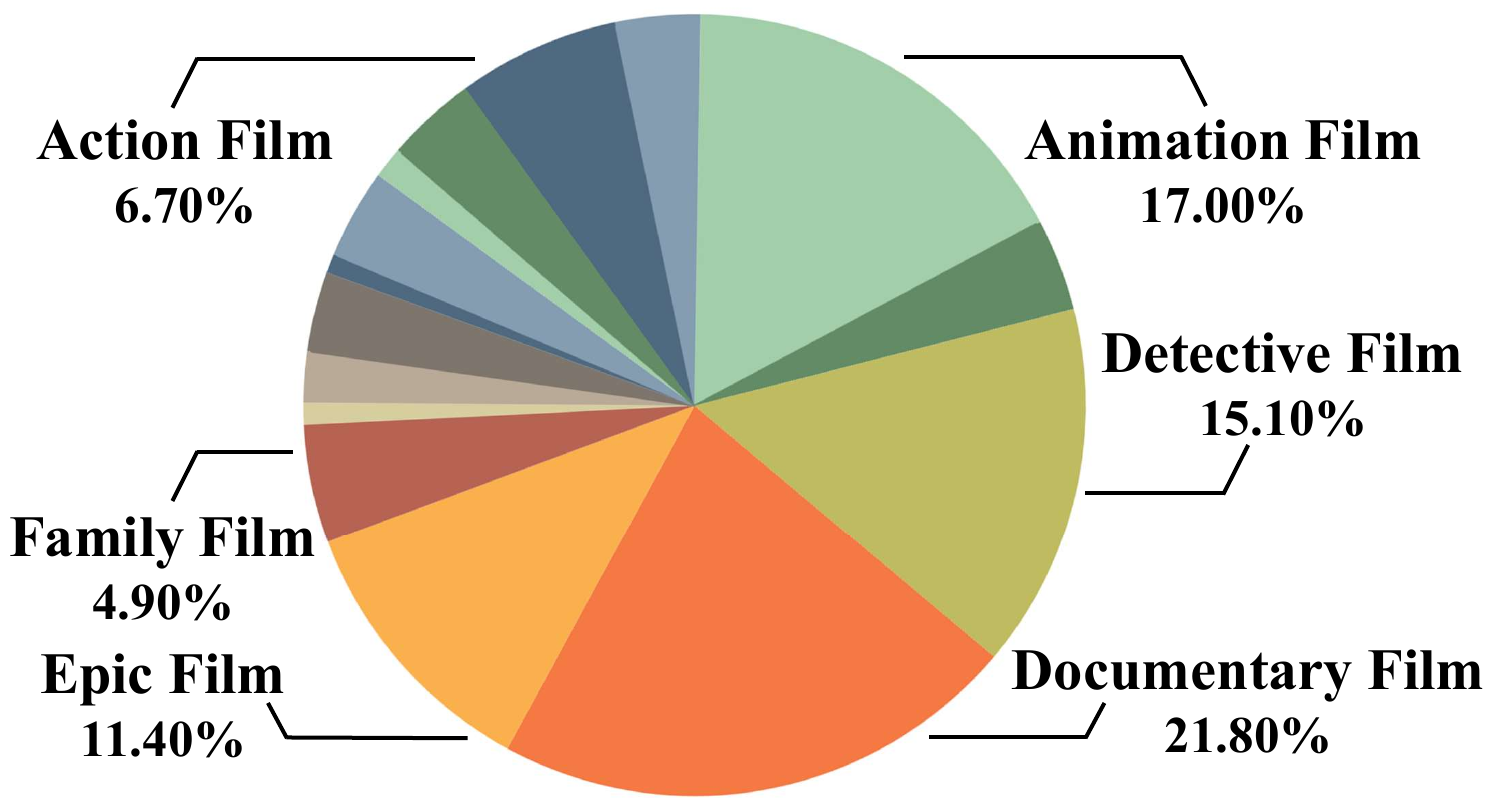}
    \caption{\textbf{Category.}}
    \label{fig:category}
  \end{subfigure}
  \begin{subfigure}{0.33\textwidth}
    \centering
    \includegraphics[width=0.95\textwidth]{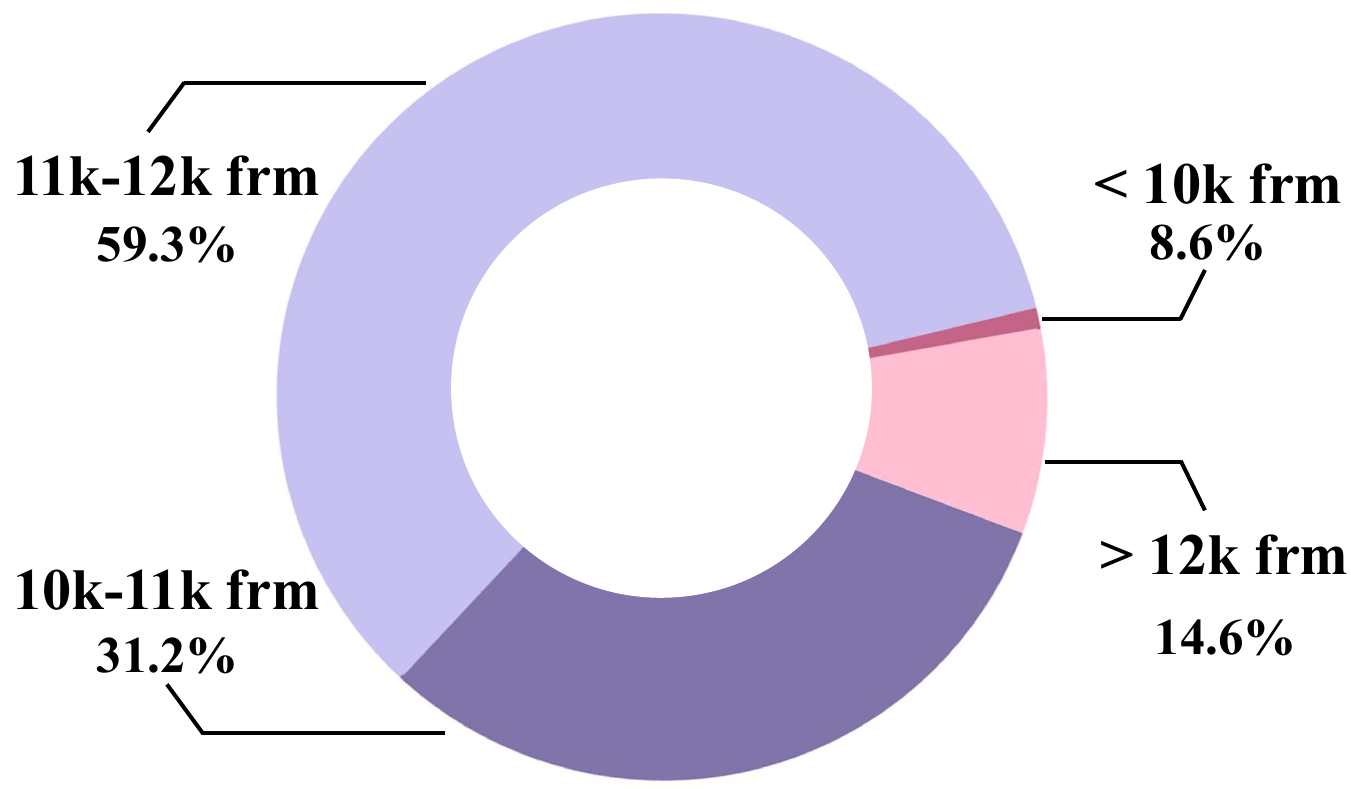}
    \caption{\textbf{Video length.}}
    \label{fig:video_length}
  \end{subfigure}
  \begin{subfigure}{0.33\textwidth}
    \centering
    \includegraphics[width=0.95\textwidth]{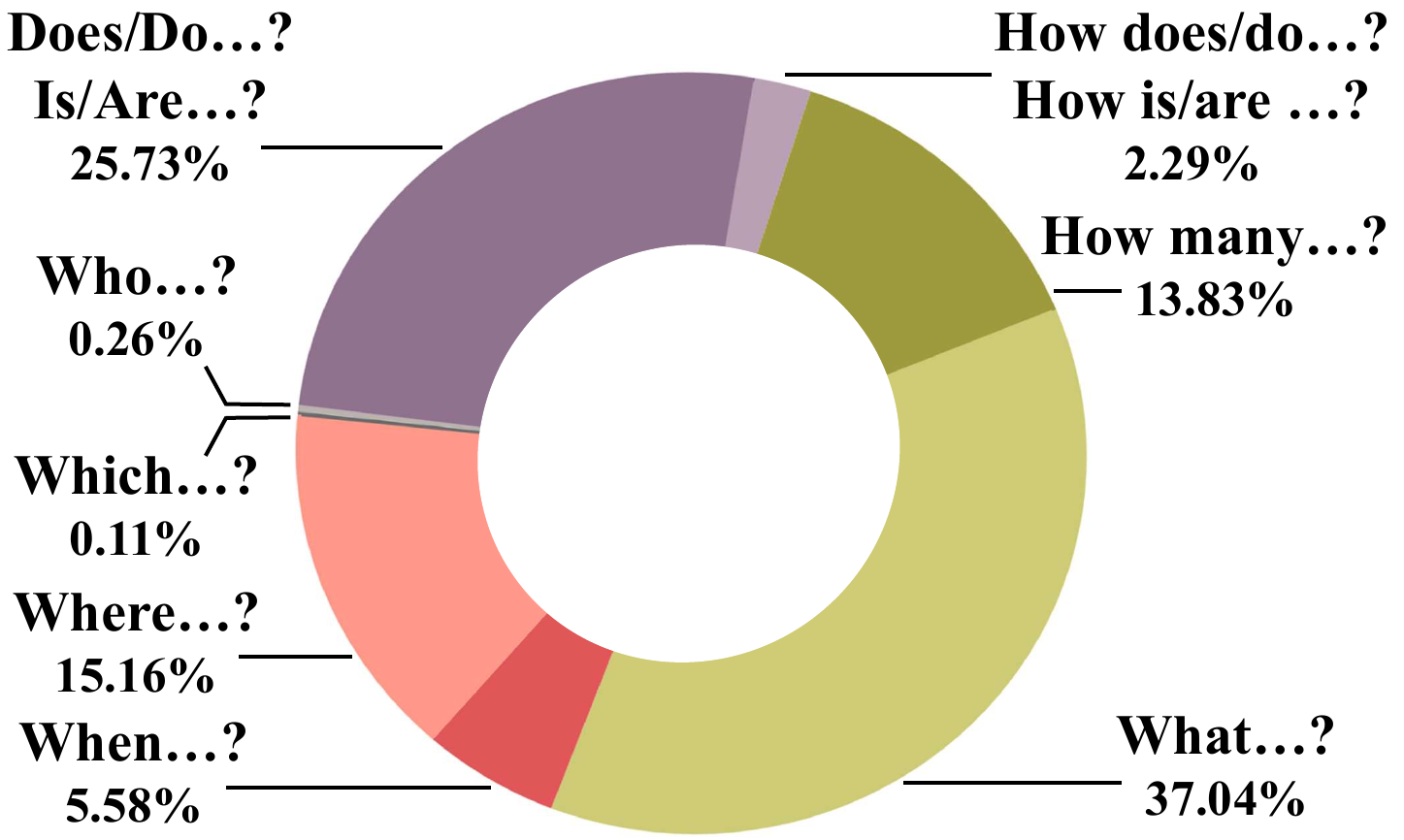}
    \caption{\textbf{Question type.}}
    \label{fig:question_type}
  \end{subfigure}
  \caption{Video-text statistics in MovieChat-1K. It encompasses a diverse set of categories, gathered from multiple question types and containing a diverse distribution of clip durations. We annotate the video categories that account for more than 4.5\% of the total (the complete list of video categories and their percentages in Appendix~\ref{sec:benchmark_stat}). ``frm" represents the number of video frames.}
  \label{fig:benchmark}
\end{figure*}

%% file: fig/wordcloud.tex
\begin{figure}[t]
    \centering
    \includegraphics[width=1\linewidth]{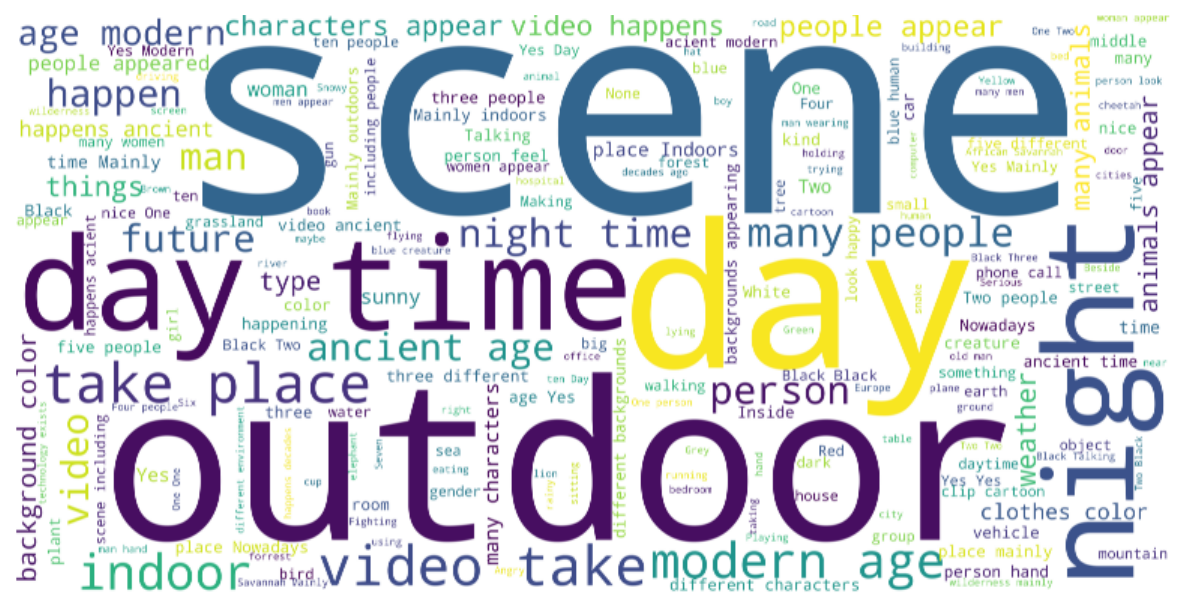}
    \caption{\textbf{Word Cloud} of the answer set in MovieChat-1K.}
    \label{fig:wordcloud}
\end{figure}

%% file: tex_arxiv/4_dataset.tex
\section{A New Benchmark: MovieChat-1K}

Previous works on building long video understanding benchmarks either focus on non-question-answering tasks~(\eg, language grounding~\cite{soldan2022mad}, generic event boundary detection~\cite{shou2021generic}, user engagement and movie metadata prediction~\cite{wu2021towards}, \etc) or lack long-form understanding evaluation~\cite{huang2020movienet}. To better evaluate the performance of MovieChat, we collect a new benchmark for long video understanding tasks, MovieChat-1K, which contains 1K high quality video clips sourced from various movies and TV series with 14K manual annotations. 

As shown in Fig.~\ref{fig:category}, we collect videos from 15 popular categories with varying distribution, including documentary film, detective film, animation film, and so on. Among these, each video comprises multiple alternating scenes, contributing to a diverse and dynamic visual narrative within the context of the collection. The visual representation in Fig.~\ref{fig:video_length} demonstrates the clip duration distribution of MovieChat-1K. Over 90\% of the videos exhibit a duration ranging from 10K to 12K frames, while 14.6\% of videos extending beyond 12K frames. Only 8.6\% of videos have duration less than 10k frames.

For each video, we manually set and provide 1 dense caption for the whole video, 3 question-answering pairs for global mode and 10 question-answering pairs with timestamps for breakpoint mode. Fig.~\ref{fig:question_type} illustrates the distribution of question types in MovieChat-1K. Note that MovieChat-1K is specifically designed for long video comprehension tasks, the majority of questions are open-ended, with only a quarter classified as multiple-choice questions, marked by initiators such as `Do,' `Does,' `Is,' or `Are.' We also compute the word distributions of our provided question-answer pairs. As illustrated in  Fig.~\ref{fig:wordcloud}, which includes common objects (people, clothes, etc.), time (day, night, etc.), scenes (indoor, outdoor, etc.), and so on.
More statistics information can be found in appendix.

%% file: tex_arxiv/5_exp.tex
\input{tab/short}
\input{tab/short_varies}
\section{Experiments}

We conduct quantitative and qualitative evaluations between MovieChat and previous methods. Additionally, we perform ablation studies to investigate MovieChat. Experimental settings and analyses can be found in appendix.

\subsection{Quantitative Evaluation}

\label{exp:quantitative}

\paragraph{Short video question-answering.} We use several widely used open-ended datasets: MSVD-QA~\cite{xu2017video}, MSRVTT-QA~\cite{xu2016msr-vtt}, and ActivityNet-QA~\cite{yu2019activitynet} for short video question-answering tasks. The evaluation process is under the assistance of LLM with the default hyper-parameter settings. The accuracy and relative scores on a scale of $0$ to $5$ are reported. Compared to previous methods~\cite{maaz2023video,li2023videochat,zhang2023llama, zhang2023video}, MovieChat achieves comparable performance even it is not specifically designed for short video question-answering tasks, as shown in Tab.~\ref{tab:short}.

\paragraph{Short video generative performance.} Following ~\cite{maaz2023video}, we employ GPT-assisted evaluation to conduct a more comprehensive comparison of the text generation performance between MovieChat and previous methods~\cite{maaz2023video,li2023videochat,yang2022zero} on processed ActivityNet-QA~\cite{yu2019activitynet}. The evaluation pipeline covers crucial metrics (including \textit{Correctness of Information}, \textit{Detailed Orientation}, \textit{Contextual Understanding}, \textit{Temporal Understanding} and \textit{Consistency}) and assigns relative scores to the generated predictions on a scale of 1-5. We present the results of the generation performance evaluation in Tab.~\ref{tab:short_varies}. The results reveal its competitive performance across all key aspects compared to previous methods. 

\paragraph{Long video question-answering.} We evaluate the long video question-answering performance of MovieChat with our proposed MovieChat-1K. We split 1,000 videos into training set~(800), test set~(100), validation set~(100) and only use test set for final performance evaluation. We select three recent LLM-based video understanding models~(\eg Video Chat~\cite{li2023videochat}, Video LLaMA~\cite{zhang2023video}, and Video-ChatGPT~\cite{maaz2023video}) as the baselines. Yet, none of those methods can support such long video~($\textgreater 10$K frames). Therefore, to accommodate their length limitations in global questions, we uniformly sample from the original video up to the maximum frame count which can be officially supported by each individual model. For breakpoint questions, we extend half of the maximum frame count before and after the breakpoint (\ie, placing the breakpoint at the center frame).

\input{tab/long}
\input{tab/long_varies}
\input{fig/ablation}

To enhance the robustness of the results, we simultaneously employ GPT-3.5~\cite{gpt3.5} and Claude~\cite{examplewebpage} as LLM assistants, with the additional support of human blind rating. We observe a discrepancy between the accuracy and relative score generated by the previously LLM-assisted evaluation method~\cite{maaz2023video} for video question-answering tasks. However, merely adjusting the prompt for the LLM cannot effectively address this issue. Therefore, after obtaining the accuracy and score from the LLM-assisted evaluation method, we implement manual filtering to remove results with inconsistent values, thus improving the reliability of our outcomes.

As shown in Tab.~\ref{tab:long}, compared to previous methods~\cite{maaz2023video,li2023videochat,zhang2023video}, MovieChat reads more video frames. In both global mode and breakpoint mode, our method maintains a performance gain in terms of the average accuracy and score provided by LLM assistants and human blind rating. We comprehensively evaluate MovieChat's question-answering performance across different question types compared to baselines. The results indicate that our approach outperforms the baselines in both open-ended and true-false questions.

\paragraph{Long video generative performance.} 

We compare the quality of answers generated by MovieChat and previous methods~\cite{maaz2023video,li2023videochat,zhang2023video} in long video question-answering on MovieChat-1K. As shown in Tab.~\ref{tab:long_varies}, with the average score provided by GPT-3.5~\cite{gpt3.5}, Claude~\cite{examplewebpage} and human bling rating, our approach continues to generate higher-quality answers even as the video contents become more extensive.

\subsection{Ablation Study}

\input{tab/videollama_score}

\input{fig/case}

\input{tab/videollama_5}

\paragraph{Short-term and long-term memory buffers.} 
As MovieChat incorporates a memory mechanism including short-term memory and long-term memory, it is imperative to evaluate how the proposed memory mechanism influences the performance. Tab.~\ref{tab:videollama_score} and Tab.~\ref{tab:videollama_5} provide the memory-dependent performance of MovieChat for long video question-answering and generative tasks with the average results of GPT-3.5~\cite{gpt3.5}, Claude~\cite{examplewebpage}, and human blind rating. MovieChat with the memory mechanism significantly outperforms the memory-independent variant, which signifies the importance of memory mechanisms.

\paragraph{Hyper-parameter ablations.} 
We perform a series of hyperparameter ablations based on the MovieChat-1K dataset to better understand MovieChat. Fig.~\ref{fig:ablation} shows the performance when ablating the length of memory buffers, consolidation length and short-term initialization with the average results of GPT-3.5~\cite{gpt3.5}, Claude~\cite{examplewebpage}, and human blind rating. The performance of MovieChat degrades when all four are significantly changed, showing the validity of our empirically chosen hyperparameyers. Fig.~\ref{fig:ablation} demonstrates that information obtained from the video expands with the growing length of memory buffers, while the loss of finer details intensifies with the fixed length of consolidation. Furthermore, using merged tokens for short-term initialization outperforms last few tokens and uniform sampling. Additionally, the length of merged tokens and the memory buffer size have a combined effect on MovieChat's performance.

\definecolor{global}{RGB}{21,96,130}
\definecolor{breakpoint}{RGB}{51,0,111}

\subsection{Case Study}

We perform an extensive case study of MovieChat on a variety of open-ended long video~(such as cartoon movie and TV series) for long video question-answering, including the \parbox[c][8pt][l]{8pt}{\colorbox{breakpoint}{}}breakpoint mode (Q\#1) and the \parbox[c][8pt][l]{8pt}{\colorbox{global}{}}global mode (Q\#2). The evaluation is conducted between MovieChat and previous methods~\cite{maaz2023video,li2023videochat,zhang2023llama} as shown in Fig.~\ref{fig:case} . For Q\#1 in breakpoint mode, we mark the timestamp when the question is asked. For long videos over $10$K frames, MovieChat is still capable of providing excellent responses to questions regarding both the current moment and the entire video content with less hallucination. More examples to show long video scene understanding and temporal understanding ability of MovieChat are available in appendix. 

%% file: tab/short.tex

\begin{table}[t]
\centering
\Large
\setlength{\tabcolsep}{8pt}
\renewcommand{\arraystretch}{1.3}
\resizebox{\linewidth}{!}{
\begin{tabular}{@{} l c c c c c c @{}}
\toprule
\multirow{2}{*}{\textbf{Method}} & \multicolumn{2}{c}{\textbf{MSVD-QA}} & \multicolumn{2}{c}{\textbf{MSRVTT-QA}} & \multicolumn{2}{c}{\textbf{ActivityNet-QA}} \\
\cline{2-7}
 & \textbf{Accuracy} & \textbf{Score} & \textbf{Accuracy} & \textbf{Score} & \textbf{Accuracy} & \textbf{Score}\\
\midrule
FrozenBiLM~\cite{yang2022zero} & 32.2 & -- & 16.8 & -- & 24.7 & -- \\
Video Chat~\cite{li2023videochat} & 56.3 & 2.8 & 45.0 & 2.5 & 26.5 & 2.2 \\
LLaMA Adapter~\cite{zhang2023llama} & 54.9 & 3.1 & 43.8 & \underline{2.7} & 34.2 & 2.7 \\
Video LLaMA~\cite{zhang2023video} & 51.6 & 2.5 & 29.6 & 1.8 & 12.4 & 1.1 \\
Video-ChatGPT~\cite{maaz2023video} & \underline{64.9} & \underline{3.3} & \underline{49.3} & \textbf{2.8} & \underline{35.2} & \underline{2.7} \\ 
\midrule
MovieChat~\textit{(Ours)} & \textbf{75.2} & \textbf{3.8} &\textbf{52.7} & 2.6 & \textbf{45.7} & \textbf{3.4}\\
\bottomrule
\end{tabular}
}
\caption{Quantitative evaluation for short video question answering with GPT-3.5~\cite{gpt3.5}. MovieChat achieves comparable performance even it is not specifically designed for for short video question-answering tasks. The best result is highlighted in bold, and the second best is underlined. }
\label{tab:short}
\end{table}

%% file: tab/short_varies.tex
\begin{table}[t]
\centering
\setlength{\tabcolsep}{8pt}
\renewcommand{\arraystretch}{0.61}
\resizebox{\linewidth}{!}{
\begin{tabular}{@{} l c c c c c c @{}}
\toprule
\scriptsize \textbf{Method} & \scriptsize \textbf{CI} & \scriptsize \textbf{DO} & \scriptsize \textbf{CU} & \scriptsize \textbf{TU} & \scriptsize \textbf{CO}\\
\midrule
 \scriptsize Video Chat~\cite{li2023videochat} &  \scriptsize 2.23& \scriptsize 2.50& \scriptsize 2.53& \scriptsize 1.94& \scriptsize 2.24\\
 \scriptsize LLaMA Adapter~\cite{zhang2023llama}& \scriptsize 2.03& \scriptsize 2.32& \scriptsize 2.30& \scriptsize 1.98& \scriptsize 2.15\\
 \scriptsize Video LLaMA~\cite{zhang2023video} & \scriptsize 1.96& \scriptsize 2.18& \scriptsize 2.16& \scriptsize 1.82& \scriptsize 1.79\\
 \scriptsize Video-ChatGPT~\cite{maaz2023video}& \scriptsize \underline{2.40}& \scriptsize \underline{2.52}& \scriptsize \underline{2.62}& \scriptsize \underline{1.98}& \scriptsize \underline{2.37}\\
\midrule
\scriptsize MovieChat~\textit{(Ours)}& \scriptsize \textbf{2.76}& \scriptsize \textbf{2.93}& \scriptsize \textbf{3.01}& \scriptsize \textbf{2.24}& \scriptsize \textbf{2.42} \\
\bottomrule
\end{tabular}
}
\caption{Quantitative evaluation for short video generation performance with GPT-3.5~\cite{gpt3.5}. CI stands for correctness of information, DO stands for detail orientation, CU stands for contextual understanding, TU stands for temporal understanding, and CO stands for consistency. The best result is highlighted in bold, and the second best is underlined.}
\label{tab:short_varies}
\end{table}

%% file: tab/long.tex
\begin{table}[t]
\centering
\setlength{\tabcolsep}{8pt}
\renewcommand{\arraystretch}{1.3}
\resizebox{\linewidth}{!}{
\begin{tabular}{@{} l c c c c c @{}}
\toprule
\multirow{2}{*}{\textbf{Method}} & \multirow{2}{*}{\textbf{\# Frames}} & \multicolumn{2}{c}{\textbf{Global Mode}} & \multicolumn{2}{c}{\textbf{Breakpoint Mode}} \\
\cline{3-6}
 & & \textbf{Accuracy} & \textbf{Score} & \textbf{Accuracy} & \textbf{Score} \\
\midrule
Video Chat~\cite{li2023videochat}& 32 & \underline{57.8} & \underline{3.00} & 46.1 & 2.29 \\
Video LLaMA~\cite{zhang2023video}& 32 & 51.7 & 2.67 & 39.1 & 2.04 \\
Video-ChatGPT~\cite{maaz2023video}& 100 & 47.6 & 2.55 & \underline{48.0} & \underline{2.45} \\ 
\midrule
MovieChat~\textit{(ours)} & 2048 & \textbf{62.3} & \textbf{3.23} & \textbf{48.3} & \textbf{2.57}\\
\bottomrule
\end{tabular}
}
\caption{Quantitative evaluation for long video question answering on MovieChat-1K test set in global mode with the average of GPT-3.5~\cite{gpt3.5}, Claude~\cite{examplewebpage} and human bling rating. HBR stands for human blind rating. The best result is highlighted in bold, and the second best is underlined.}
\label{tab:long}
\end{table}

%% file: tab/long_varies.tex

\begin{table}[t]
\centering
\setlength{\tabcolsep}{8pt}
\renewcommand{\arraystretch}{0.61}
\resizebox{\linewidth}{!}{
\begin{tabular}{@{} l c c c c c c @{}}
\toprule
\scriptsize \textbf{Method} & \scriptsize \textbf{CI} & \scriptsize \textbf{DO} & \scriptsize \textbf{CU} & \scriptsize \textbf{TU} & \scriptsize \textbf{CO}\\
\midrule
\scriptsize Video Chat~\cite{li2023videochat} & \scriptsize \underline{3.04} & \scriptsize \underline{2.75} & \scriptsize \underline{3.09} & \scriptsize \underline{3.00}& \scriptsize  \underline{3.21}\\
\scriptsize Video LLaMA~\cite{zhang2023video} & \scriptsize 2.75 & \scriptsize 2.24& \scriptsize 2.83& \scriptsize 2.62& \scriptsize 2.97\\
\scriptsize Video-ChatGPT~\cite{maaz2023video}& \scriptsize 2.37& \scriptsize 2.30& \scriptsize 2.58& \scriptsize 2.49& \scriptsize 2.69\\
\midrule
\scriptsize MovieChat~\textit{(Ours)}& \scriptsize \textbf{3.11}& \scriptsize \textbf{2.93} & \scriptsize \textbf{3.24}& \scriptsize \textbf{3.17}& \scriptsize \textbf{3.25} \\
\bottomrule
\end{tabular}
}
\caption{Quantitative evaluation for long video generation performance in global mode with the average of GPT-3.5~\cite{gpt3.5}, Claude~\cite{examplewebpage} and human blind rating. CI stands for correctness of information, DO stands for detail orientation, CU stands for contextual understanding, TU stands for temporal understanding, and CO stands for consistency. The best result is in bold, and the second best is underlined.}
\label{tab:long_varies}
\end{table}

%% file: fig/ablation.tex

\begin{figure*}[t]
  \centering
  \includegraphics[width=\textwidth]{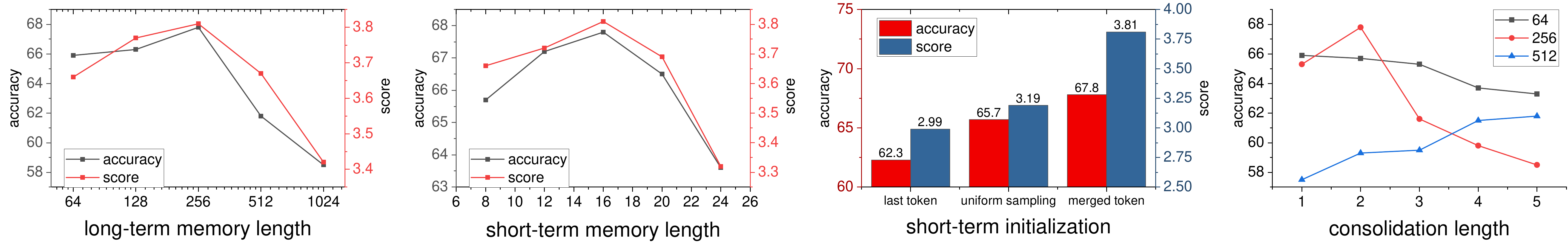}
  \caption{Hyperparameter ablation studies on how length of long-term memory buffer $l_{length}$, short-term memory buffer $l_{short}$, consolidation length $l_{merge}$ and short-term initialization affect the performance of MovieChat on long video understanding. We set $l_{short}=16$, $l_{merge}=2$ in ablation study of long-term memory,  $l_{long}=256$, $l_{merge}=2$ in ablation study of short-term memory and $l_{short}=16$ in ablation study of consolidation length and short-term initialization. }
  \label{fig:ablation}
\end{figure*}



%% file: tab/videollama_score.tex
\begin{table}[t]
\centering
\setlength{\tabcolsep}{12pt}
\renewcommand{\arraystretch}{0.7}
\resizebox{\linewidth}{!}{
\begin{tabular}{@{} c c c c c @{}}
\toprule
\multirow{2}{*}{\textbf{\scriptsize Method}} & \multicolumn{2}{c}{\textbf{\scriptsize Global Mode}} & \multicolumn{2}{c}{\textbf{\scriptsize Breakpoint Mode}} \\
\cline{2-5}
 & \scriptsize {Accuracy} & \scriptsize {Score} & \scriptsize {Accuracy} &  \scriptsize {Score}\\
\midrule
 \rule{0pt}{5pt} \scriptsize w/o MM &   \scriptsize 51.4&  \scriptsize 3.10&  \scriptsize 38.2&  \scriptsize 2.31 \\
  \rule{0pt}{5pt} \scriptsize base&  \scriptsize \textbf{67.8}&  \scriptsize \textbf{3.81}&  \scriptsize \textbf{50.4}& \scriptsize \textbf{2.96} \\
\bottomrule
\end{tabular}
}
\caption{Ablation study on how memory mechanism (MM) affects the long video question answering. The best result is in bold.}
\label{tab:videollama_score}
\vspace{-5pt}
\end{table}

%% file: fig/case.tex
\definecolor{global}{RGB}{21,96,130}
\definecolor{breakpoint}{RGB}{51,0,111}
\definecolor{hall}{RGB}{183,165,122}

\begin{figure*}[t]
	\centering
    \includegraphics[width=0.95\linewidth]{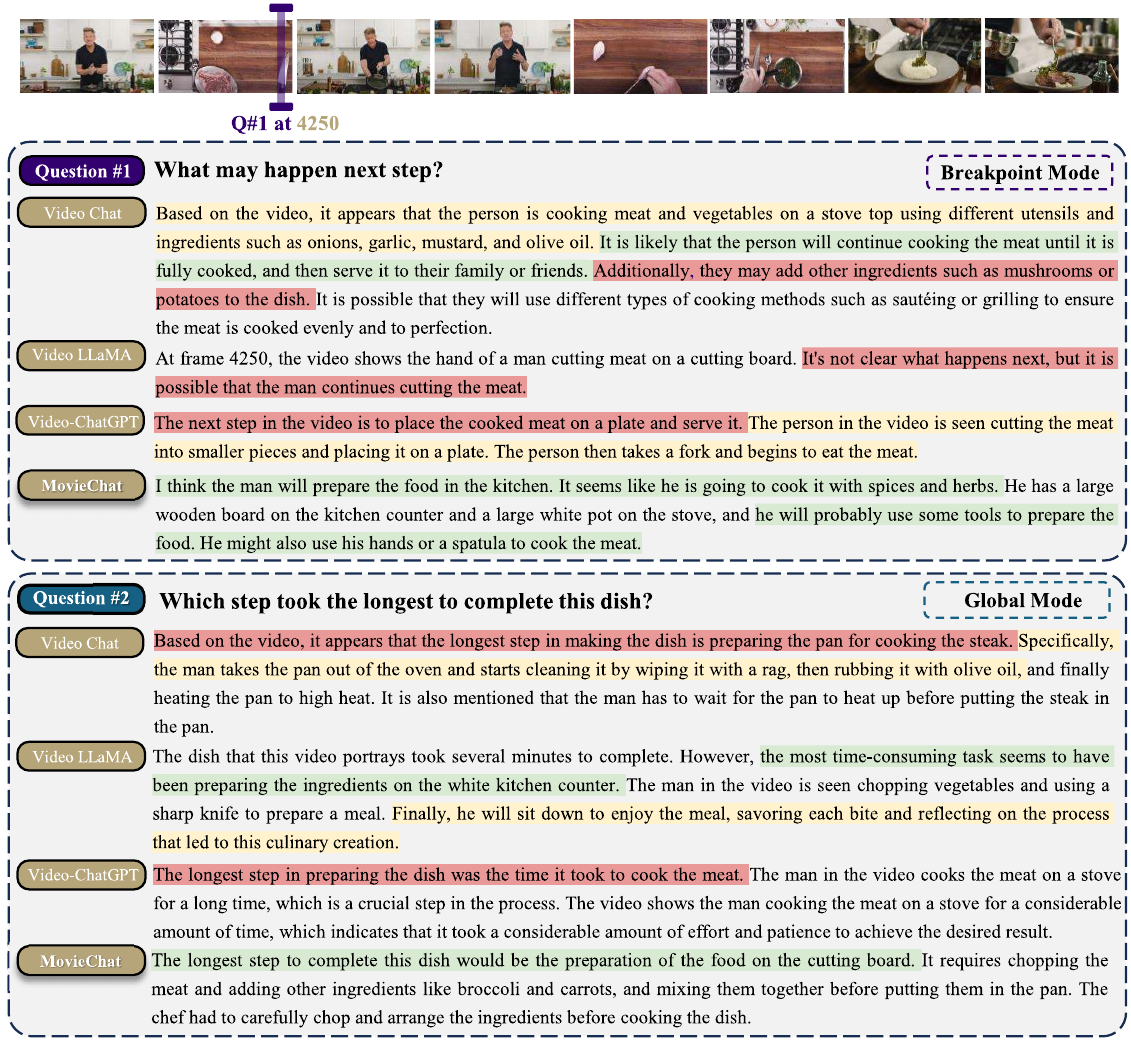}
	\caption{Question and answer about a clip from \textit{YouTube}, which is a tutorial on how to cook steak. The entire instructional process begins with marinating the steak, followed by pan-searing it, preparing side dishes, and ultimately plating the meal. \colorbox[RGB]{217,234,211}{Green} ( \colorbox[RGB]{234,153,153}{Red} ) highlights the correct (wrong) answer and \colorbox[RGB]{255,242,204}{yellow} indicates that the model is hallucinating.}
 \vspace{20pt}
\label{fig:case}
\end{figure*}

%% file: tab/videollama_5.tex
\begin{table}[t]
\centering
\setlength{\tabcolsep}{8pt}
\renewcommand{\arraystretch}{1.3}
\resizebox{\linewidth}{!}{
\begin{tabular}{@{} c c c c c c c c c c c @{}}
\toprule
\multirow{2}{*}{\textbf{Method}} & \multicolumn{5}{c}{\textbf{Global Mode}} & \multicolumn{5}{c}{\textbf{Breakpoint Mode}} \\
\cline{2-11}
& \textbf{CI} & \textbf{DO} & \textbf{CU} & \textbf{TU} & \textbf{CO} & \textbf{CI} & \textbf{DO} & \textbf{CU} & \textbf{TU} & \textbf{CO}\\
\midrule
w/o MM &  3.30&  2.53&  3.28&  2.77& 3.42& 2.42& 2.85& 2.87& 2.00& 2.87 \\
base&  \textbf{3.32}&  \textbf{3.28}&  \textbf{3.40}&  \textbf{2.97}& \textbf{3.48}& \textbf{2.97}& \textbf{3.24}& \textbf{3.31}& \textbf{2.70}& \textbf{3.45}\\
\bottomrule
\end{tabular}
}
\caption{Ablation study on how memory mechanism (MM) affects the long video generative performance. CI stands for correctness of information, DO stands for detail orientation, CU stands for contextual understanding, TU stands for temporal understanding, and CO stands for consistency. The best result is in bold.}
\label{tab:videollama_5}
\vspace{-5pt}
\end{table}

%% file: tex_arxiv/6_conclusion.tex
\section{Limitation}

Although MovieChat has demonstrated impressive abilities in long video understanding, it is still an early-stage prototype and has some limitations, including: 1) Limited perception capacities. MovieChat’s performance is hindered by the pretrained short video understanding model. 2) Inadequate Time Processing. MovieChat provides only rough estimates of the duration proportions of events within long videos, lacking precision in temporal details.

\section{Conclusion}

Conclusively, we presents an innovative video understanding system integrating video foundation models and large language models. By incorporating a memory mechanism represented by tokens in Transformers, our proposed system, MovieChat overcomes challenges associated with analyzing long videos.
MovieChat achieves state-of-the-art performance in long video understanding, surpassing existing systems limited to handling videos with few frames. 

%% file: tex_arxiv/7_appendix.tex
\appendix
\renewcommand\thefigure{\Alph{section}\arabic{figure}}
\renewcommand\thetable{\Alph{section}\arabic{table}}
\setcounter{figure}{0}
\setcounter{table}{0}

The supplementary material is structured as follows:

\begin{enumerate}
\item We first present schematic diagram of the memory consolidation algorithm of MovieChat in Section~\ref{sec:merge}.

\item We provide detailed supplementary statistical information for MovieChat-1K in Section~\ref{sec:benchmark_stat}.

\item The prompt template we use for LLM-Assisted Evaluation is shown in Section~\ref{sec:gpt-eval}.

\item We also list the hyperparameter settings of MovieChat in Section~\ref{sec:hyper-param}.

\item We mention the specifical LLM-Assisted Evaluation method employed for the assessment of short video generative performance in Section~\ref{sec:gpt-eval-varies}.

\item In comparison to the LLM currently used by MovieChat, we switch to a different LLM and compare the results in Section~\ref{sec:llm}.

\item To avoid the impact of misjudgments by LLM assistants on the results, we introduce the manual filtering strategy in Section~\ref{sec:filter}.

\item The performance of MovieChat varies across different categories of questions, and we present the results in Section~\ref{sec:type_result}.

\item We add the result of quantitative evaluation for long video generative performance in breakpoint mode in Section~\ref{sec:break_quan}.

\item To demonstrate the outstanding performance of MovieChat across a wide range of categories, we calculate the Pearson correlation coefficient of different score methods in Section~\ref{sec:pear}.

\item We then list the evaluaotion results with GPT-3.5~\cite{gpt3.5}, Claude~\cite{examplewebpage} and human blind rating in Section~\ref{sec:eva_methods}.

\item We conduct result analyses for the ablation study on hyperparameter settings in Section~\ref{sec:hyp_ana}.

\item Lastly, we give more examples for scene understanding and temporal understanding of MovieChat in Section~\ref{sec:task_case}.
  
\end{enumerate}

\section{Memory consolidation algorithm of MovieChat.}
\label{sec:merge}
\input{fig/merge}

As shown in Fig.~\ref{fig:merge}, for each sampled frame $x_{i}$, we calculate its similarity with adjacent frames. After that, we select the pair with the greatest similarity, merge and replace these two frames, resulting in a new sequence. We conduct the merge operation repeatedly until the count of existing frames in short-term memory reaches the predefined value.

\section{MovieChat-1K Statistics Information}
\label{sec:benchmark_stat}

\paragraph{Distribution of video categories.}
MovieChat-1K contains videos from 15 popular categories with varying distribution. As shown in Tab.~\ref{tab:cate}, every video comprises multiple alternating scenes.
\input{tab/cate}

\paragraph{Video information and visual question-answer data format.}

To the best of our knowledge, a long video understanding dataset has not
yet been established. Our work represents the initial step in creating and making it publicly available.We create MovieChat1K, containing 1k long videos and corresponding 1k dense captions, and 13k visual question-answer pairs.One visual example of these arrangements is provided in Figure~\ref{fig:data_format}.
\input{fig/data_format}

\input{fig/question}

\input{fig/answer}

\paragraph{Sentence length distribution of question-answer pairs.}
MovieChat1K exhibits diverse lengths of question-answer pairs in the segmented clip level. Fig.~\ref{fig:question} and Fig.~\ref{fig:answer} demonstrate the length distribution of question-answer pairs in different modes. Despite the distribution of question-answer pairs varies between the global mode and breakpoint mode, the majority of questions tends to concentrate between 5-15 words in length, while the length of answers generally have fewer than 10 words.

\paragraph{Stastics information of dense captions.}
\input{fig/caption_length}
\input{fig/caption_wordcloud}

To facilitate a more detailed understanding of long videos, we provide a dense caption for each video. As shown in Fig.~\ref{fig:caption_length}, MovieChat-1K exhibits diverse caption lengths in the segmented clip level. Approximately two-thirds
of the clips have captions with 100-149 words, while one-fifth of the clip captions have fewer than 100 words. About 11\% of clips have long captions with more than 150 words.

To analyze the word distribution of our generated
captions, we compute their distributions. The resulting word
distribution of the captions is presented in Fig.~\ref{fig:caption_wordcloud}, which includes common objects (man, woman, people,
girl, etc.), attributes (detective, various, small, white, etc.), locations (inside, behind, south, next, etc.),
scenes (room, house, building, office, etc.), actions/events (talk, enter, leave, take, etc.), and
more.

In terms of actionness, MovieChat-1K captions contains nearly the same number of verbs as with the WebVid10M dataset~\cite{DBLP:journals/corr/abs-2104-00650}. To evaluate this, we use the NLTK toolkit to analyze the number of verbs in captions, focusing on extracting and tagging all unique verbs. We find a total of 109,485 verbs in the WebVid10M caption dataset, while the MovieChat-1K captions contain 102,988 unique instances of verbs. While these counts may not be entirely accurate due to our simple counting method, we believe they provide a rough indication of the actionness of the two datasets.

\paragraph{Comparison between MovieChat-1K and other benchmarks.}
MovieChat-1K provides a large-scale benchmark for long video understanding, which contains
1K movies, 1K dense captions and 13k question-answer pairs. The comparison between different
datasets are shown in Tab.~\ref{tab:benchmarks}. 
It is evident that MovieChat-1K provides the longest average duration for movie clips. MovieQA~\cite{tapaswi2016movieqa} exclusively offers question-answer pairs related to movies, while MovieGraphs~\cite{vicol2018moviegraphs} supplies captions associated with movies. Unlike other datasets, MovieNet~\cite{huang2020movienet} encompasses three main types of texts: subtitle, synopsis, and script, excluding question-answer pairs. Additionally, the synopsis category is designed for the entire movie rather than video clips. Consequently, MovieChat-1K is more suitable for studying long video comprehension compared to other datasets.
\input{tab/benchmarks}

\section{LLM-Assisted Evaluation for the short video question-answering task.}
\label{sec:gpt-eval}
Following~\cite{maaz2023video}, we use LLM-Assisted Evaluation for the short video question-answering task in Section~\ref{exp:quantitative}. Given the question, correct answer, and predicted answer by the model, the LLM assistants should return the \textit{True} or \textit{False} judgement and relative score ($0$ to $5$). The whole prompt is shown in Fig.~\ref{fig:prompt}. It takes about $250$ tokens per question. We report the baseline results of short video question-answering from \url{https://github.com/mbzuai-oryx/Video-ChatGPT}.
\setcounter{figure}{0}
\input{fig/prompt}

\section{Hyperparameter Setting}
\label{sec:hyper-param}
\input{tab/hyper-param}

We report the detailed hyperparameter settings of MovieChat in Tab.~\ref{tab:hyper}. The sliding window size of MovieChat is set to 16, which means that every slide involves the extraction of 16 frames. We configure the short-term memory to consist of 18 frames, with each frame containing 32 tokens. When the short-term memory reaches its capacity, it is directed to the memory consolidation module to be merged into 2 representative frames. The 2 frames are simultaneously input into the long-term memory with a total length of 256 and used to reinitialize the short-term memory.

\section{LLM-Assisted Evaluation for short video generative performance.}
\label{sec:gpt-eval-varies}
We use LLM-Assisted Evaluation proposed by~\cite{maaz2023video} for short video generative performance in Section~\ref{exp:quantitative}. The evaluation pipeline assesses various capabilities of the model and assigns a relative score ($1$ to $5$) to the generated predictions, in the following five aspects:\textit{ Correctness of Information}, \textit{Detail Orientation}, \textit{Contextual Understanding}, \textit{Temporal Understanding} and \textit{Consistency}. We follow the corresponding prompts provided in \url{https://github.com/mbzuai-oryx/Video-ChatGPT}
and report the baseline results of short video generative performance from it.

\section{Ablation study on large language models.}
\label{sec:llm}
\input{tab/llm_score}
\input{tab/llm_5}

\input{fig/llm}

Most previous video understanding methods~\cite{maaz2023video,li2023videochat,zhang2023llama, zhang2023video} primarily employed LLama~\cite{touvron2023llama} and its variants~\cite{stablevicuna-github} as text decoders. With the average results of GPT-3.5~\cite{gpt3.5}, Claude~\cite{examplewebpage} and human blind rating, Tab.~\ref{tab:llm_score} and Tab.~\ref{tab:llm_5} illustrate how the performance of MovieChat changes when using LLama~\cite{touvron2023llama} and LLama2~\cite{touvron2023llama2} as the large language model respectively. 

Contrary to our hypothesis, under every evaluation conditions, the performance metrics of MovieChat with LLama2~\cite{touvron2023llama2} hardly surpassed those of MovieChat with LLama~\cite{touvron2023llama}. We further investigate a specific example to analyze this phenomenon. As shown in Fig.~\ref{fig:llm}, the bold segments represent direct responses to the questions from two versions of MovieChat. MovieChat with LLama~\cite{touvron2023llama} provided answers that are more aligned with the video content. Surprisingly, MovieChat with LLama2~\cite{touvron2023llama2} offer an approximation of the time required for each step (indicated by underlines Fig.~\ref{fig:llm}). While its time estimates do not precisely match the actual durations, the proportion of time provided was realistic. Even though LLama2~\cite{touvron2023llama2} cannot obtain specific time information when processing feature-rich video frames, MovieChat's memory buffer design allows for dense sampling of video frames, enabling LLama2~\cite{touvron2023llama2} to estimate the proportion of time for each scene based on adjacent similar frames. Therefore, we propose that the lower evaluation metric results of MovieChat with LLama2~\cite{touvron2023llama2} compared to MovieChat with LLama~\cite{touvron2023llama} may be attributed to the question-answer pairs provided by the dataset.

\section{Manual filtering strategy for LLM-Assisted Evaluation.}
\label{sec:filter}
For each test data, ~\cite{maaz2023video} utilized GPT-3.5~\cite{gpt3.5} to provide an evaluation result in terms of a 'yes/no' response and a corresponding score, as demonstrated in Fig.~\ref{fig:prompt}. The score is an integer value ranging from 0 to 5, where a score of 5 indicates the highest degree of meaningful correspondence. However, we observe instances where GPT-3.5~\cite{gpt3.5} offered judgments and scores that do not align, such as providing a 'yes' response with a score of 0 or a 'no' response with a score of 5. This discrepancy has the potential to impact the accuracy of results and introduce fluctuations. We adapt the prompts used for GPT-3.5~\cite{gpt3.5} with the aim of addressing this concern and did not yield the desired mitigation.
Hence, we introduce an artificial filtering strategy. For each evaluation result generated by GPT-3.5~\cite{gpt3.5}, we conduct manual screening. We retain only those outcomes that exhibited consistency between the 'yes/no' judgments and the associated scores, thus enhancing the reliability of the evaluations. 
Similarly, we applied the same filtering strategy to the evaluation results generated by Claude~\cite{examplewebpage}.

\section{Quantitative evaluation for long video different types question answering.}
\label{sec:type_result}
\input{tab/type_result_global}

\input{tab/type_result_break}

As shown in Fig.~\ref{fig:benchmark}, MovieChat-1K contains question-answer pairs of varies types. To better assess the performance of MovieChat, we conduct evaluations on the long video question answering task using various types of questions. We roughly categorize the question types into multiple-choice questions and open-ended questions. With the average results of GPT-3.5~\cite{gpt3.5}, Claude~\cite{examplewebpage} and human blind rating, Tab.~\ref{tab:type_result_global} and Tab.~\ref{tab:type_result_break} respectively present the accuracy and scores of MovieChat and the baseline across different question categories in both global mode and breakpoint mode. In various research conditions, our approach consistently outperforms the baseline, thus substantiating the robustness of MovieChat.

\section{Quantitative evaluation for long video generative performance in breakpoint mode}
\label{sec:break_quan}
\input{tab/break_5}

Similar to Tab.~\ref{tab:long_varies}, with the average results of GPT-3.5~\cite{gpt3.5}, Claude~\cite{examplewebpage} and human blind rating, Tab.~\ref{tab:break_5} demonstrates that our method outperforms the baseline in long video generative performance in breakpoint mode.

\section{Pearson correlation coefficient of different score
methods.}
\label{sec:pear}
\input{fig/pear}
\input{tab/Pearson}

The Pearson correlation coefficient is represented by the formula:

$$r_{xy}=\frac{\sum_{i=1}^n(x_i-\bar{x})(y_i-\bar{y})}{\sqrt{\sum_{i=1}^n(x_i-\bar{x})^2}\sqrt{\sum_{i=1}^n(y_i-\bar{y})^2}}$$

where $r_{xy}$ is the Pearson correlation coefficient between two variables $x$ and $y$, $x_{i}$ and $y_{i}$ are the individual sample points for variables $x$ and $y$, $\overline{x}$ and $\overline{y}$ are the averages of the $x$ and $y$ samples respectively, and $n$ is the number of sample points.
The formula essentially assesses the extent of linear correlation between two variables by evaluating the product of their deviations from their respective means. The numerator represents the covariance between the two variables, and the denominator normalizes this value, ensuring that the coefficient remains between -1 and +1. The Pearson correlation coefficient quantifies the extent to which two variables co-vary in comparison to their individual variations.

As shown in Tab.~\ref{tab:pearson}and Fig.~\ref{fig:pearson}, we conduct pearson correlation analysis between GPT-3.5~\cite{gpt3.5}, Claude~\cite{examplewebpage}, and human blind rating. The result indicates a substantial agreement among these evaluation methods. The alignment of scores across different score methods strengthens the reliability of our assessment. Crucially, our proposed method, MovieChat outperforms previous methods~\cite{maaz2023video,li2023videochat,zhang2023llama, zhang2023video} in long video understanding tasks. The superior performance of MovieChat is evident across a broad spectrum of categories, suggesting that our model not only has a deeper understanding of long videos and respective questions but also exhibits a more accurate and consistent ability to generate relevant responses.

\section{Evaluation results with GPT, Claude and human blind rating.}
\label{sec:eva_methods}

As shown in~\ref{tab:long_gpt}--\ref{tab:break_5_human}, we provide detailed scoring results for GPT-3.5~\cite{gpt3.5}, Claude~\cite{examplewebpage}, and human blind rating across various experiments.

\input{tab/long_gpt}
\input{tab/long_claude}
\input{tab/long_human}

\input{tab/long_varies_gpt}
\input{tab/long_varies_claude}
\input{tab/long_varies_human}

\input{tab/break_5_gpt}

\input{tab/break_5_claude}

\input{tab/break_5_human}

\section{Analysis on hyperparameter ablations.}
\label{sec:hyp_ana}
As the lengths of the short-term and long-term memory buffers increase, the information acquired by MovieChat from the video expands, as illustrated in Fig.~\ref{fig:ablation}. However, more video compression leads to the loss of more detailed information, while the length of the merged tokens remains constant. Therefore, as the lengths of two memory buffers increase, the performance of MovieChat exhibits a trend of initially rising and then declining.

Fig.~\ref{fig:ablation} demonstrates how memory consolidation influences the performance. Since the LLM-based evaluation shows a positive correlation between accuracy and score, we use accuracy to gauge performance. When memory buffer parameters remain constant, shorter merged tokens indicate increased frame information compression, potentially resulting in information loss when excessive. Conversely, longer merged tokens, despite retaining a greater extent of short-term memory in the face of compression, correspondingly result in less overall information acquisition. Moreover, when the length of the memory buffer changes, as exemplified by long-term memory, the corresponding peak performance of MovieChat shifts in response to the altered length of merged tokens. This demonstrates the need to strike a balance between dense information extraction and information compression in long video understanding tasks.

We also conduct experiments to compare various methods for initializing the short-term memory, including selecting the last few tokens, uniform sampling, and using merged tokens. The results indicate that the use of merged tokens produces the best performance. When initializing the next short-term memory with the last few tokens from the previous short-term memory, it is unable to adequately represent the information from the previous time step. Consequently, this leads to the final merged tokens being either repetitive or lacking coherence with the previous time step. Uniform sampling faces similar issues, but it manages to capture information with representative frames from the previous time step. Consequently, its performance surpasses that of initializing with the last few tokens, yet it remains inferior to using merged tokens for initialization.

\section{Examples for scene understanding and temporal understanding of MovieChat.}
\label{sec:task_case}

We perform an extensive case study of MovieChat on a variety of open-ended long video~(such as cartoon movie in and TV series) for long video question-answering and captioning task, including the \parbox[c][8pt][l]{8pt}{\colorbox{global}{}}global mode and the \parbox[c][8pt][l]{8pt}{\colorbox{breakpoint}{}}breakpoint mode. 
The evaluation tasks include scene understanding and temporal understanding as shown in Fig.~\ref{fig:case1}, Fig.~\ref{fig:case2}, Fig.~\ref{fig:case3} and Fig.~\ref{fig:case4}. For Q\#1 and Q\#2, we remarks timestamps in frames. For long videos over $10$K frames, MovieChat is still capable of providing excellent responses to questions regarding both the current moment and the entire video content. 

\input{fig/case1}
\input{fig/case2}

\clearpage

%% file: fig/merge.tex
\definecolor{global}{RGB}{21,96,130}
\definecolor{breakpoint}{RGB}{51,0,111}
\definecolor{hall}{RGB}{183,165,122}

\begin{figure*}[!t]
	\centering
    \includegraphics[width=0.8\linewidth]{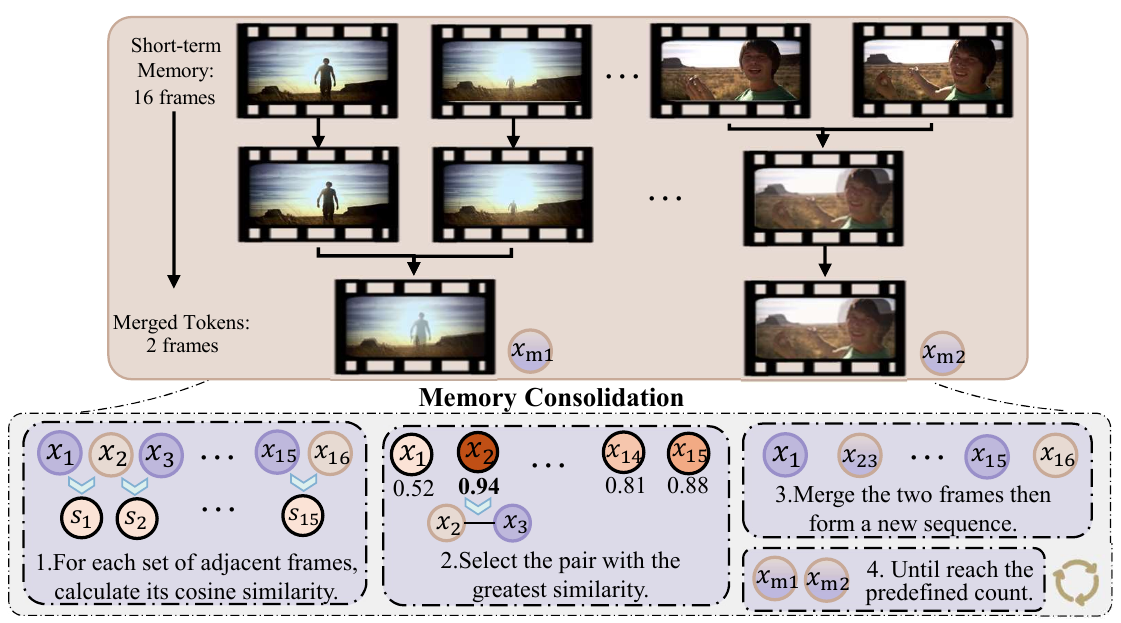}
	\caption{Question and answer about clips from \textit{YouTube}, which is a tutorial on how to cook steak. The entire instructional process begins with marinating the steak, followed by pan-searing it, preparing side dishes, and ultimately plating the meal.}
 \vspace{20pt}
\label{fig:merge}
\end{figure*}

%% file: tab/cate.tex
\begin{table}[h]
\centering
\setlength{\tabcolsep}{6pt}
\renewcommand{\arraystretch}{0.7}
\resizebox{0.6\linewidth}{!}{
\begin{tabular}{ c | c}
\toprule
\textbf{\tiny Category} & \textbf{\tiny Percentage} \\
\midrule
\tiny Documentary Film & \tiny 21.80\% \\
\tiny Animation Film & \tiny 17.00\% \\
\tiny Detective Film & \tiny 15.10\% \\
\tiny Epic Film & \tiny 11.40\% \\
\tiny Action Film & \tiny 6.70\% \\
\tiny Family Film & \tiny 4.90\% \\
\tiny Crime Film & \tiny 3.80\% \\
\tiny Science Fiction Film & \tiny 3.70\% \\
\tiny War Film & \tiny 3.70\% \\
\tiny Adventure Film & \tiny 3.50\% \\
\tiny Romance Film & \tiny 3.30\% \\
\tiny History Film & \tiny 2.10\% \\
\tiny Suspense Film & \tiny 1.30\% \\
\tiny Fantasy & \tiny 0.90\% \\
\tiny School Film & \tiny 0.80\% \\
\bottomrule
\end{tabular}
}
\caption{Distribution of video categories in MovieChat-1K.}
\label{tab:cate}
\end{table}

%% file: fig/data_format.tex
\begin{figure*}[h]
    \centering
    \includegraphics[width=1\linewidth]{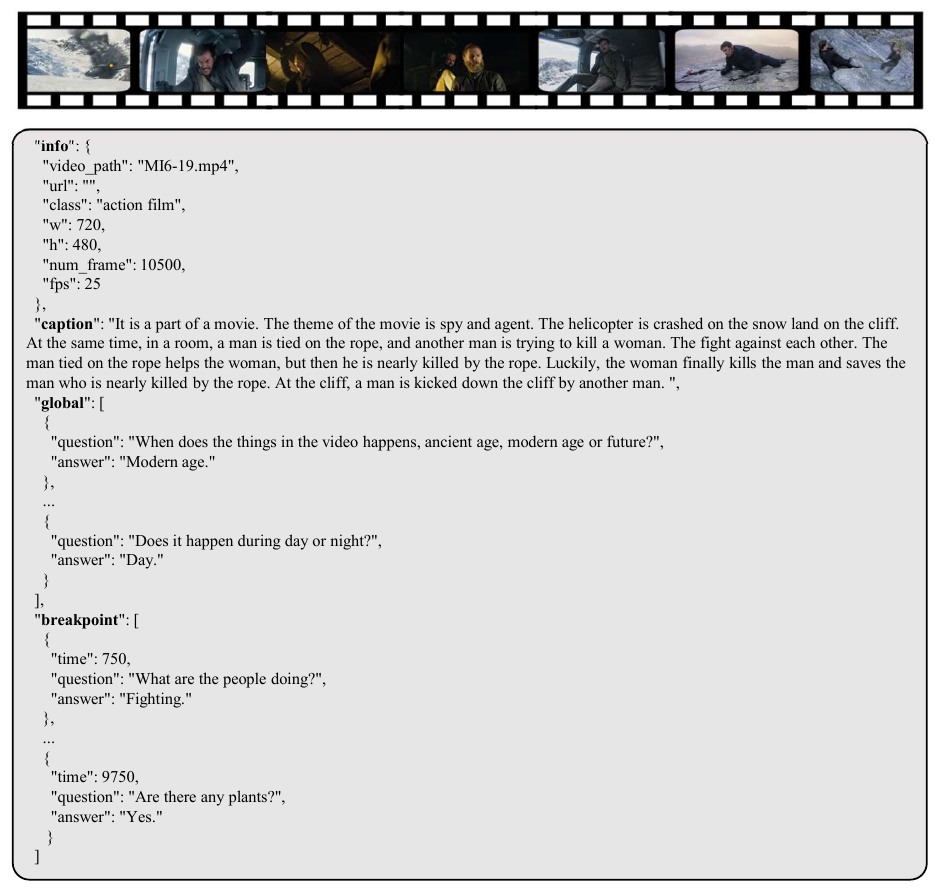}
    \caption{Video information and visual question-answer data format in MovieChat1K. }
    \label{fig:data_format}
\end{figure*}

%% file: fig/question.tex
\begin{figure}[t]
    \centering
    \begin{subfigure}[b]{0.4\textwidth}
    \includegraphics[width=1\linewidth]{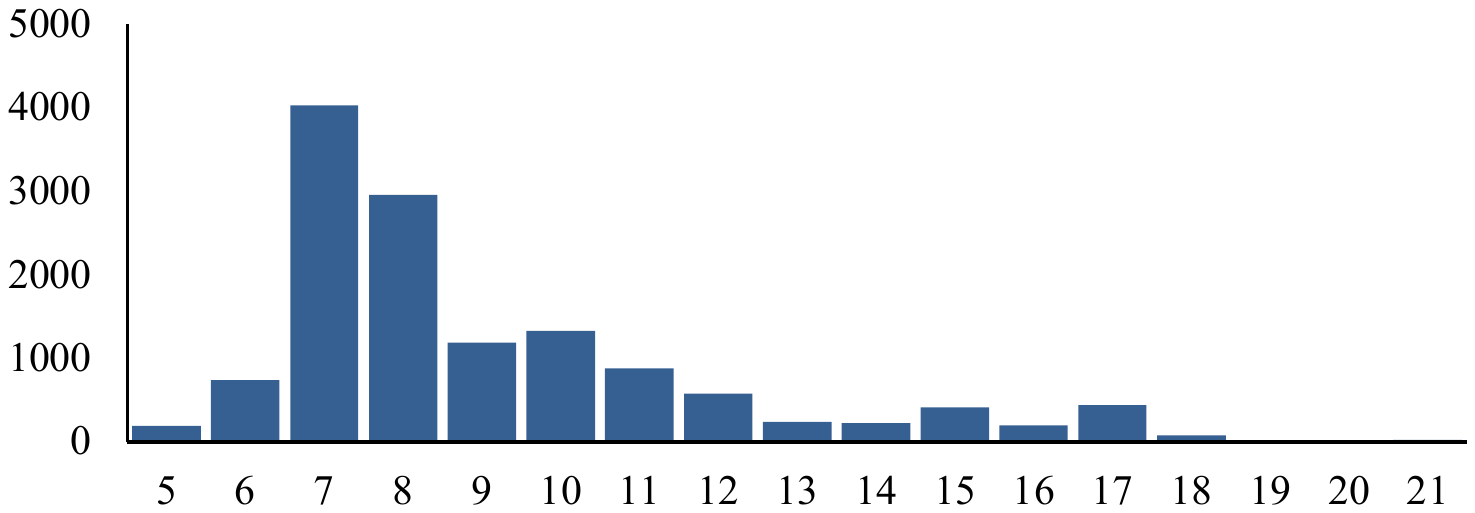}
    \vspace{-45pt}
    \caption{Length of total questions.}
    \label{fig:subfig-a}
  \end{subfigure}
  \begin{subfigure}[b]{0.4\textwidth}
    \vspace{-30pt}
    \includegraphics[width=1\linewidth]{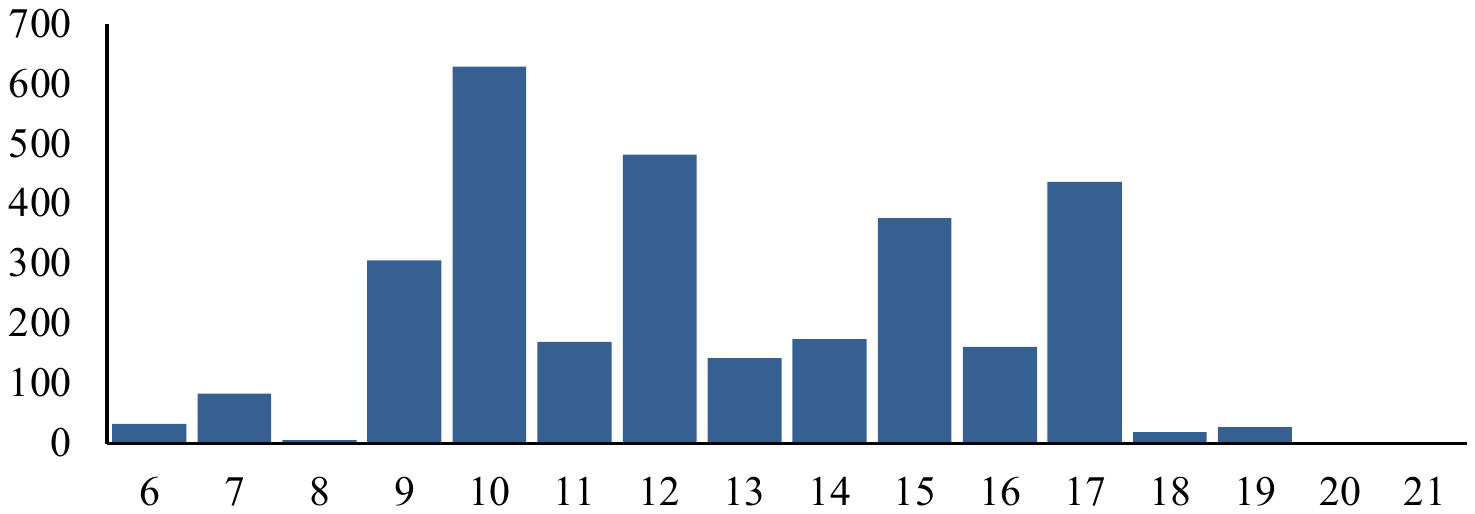}
    \vspace{-45pt}
    \caption{Length of global questions.}
    \label{fig:subfig-b}
  \end{subfigure}
  \begin{subfigure}[b]{0.4\textwidth}
    \vspace{-30pt}
    \includegraphics[width=1\linewidth]{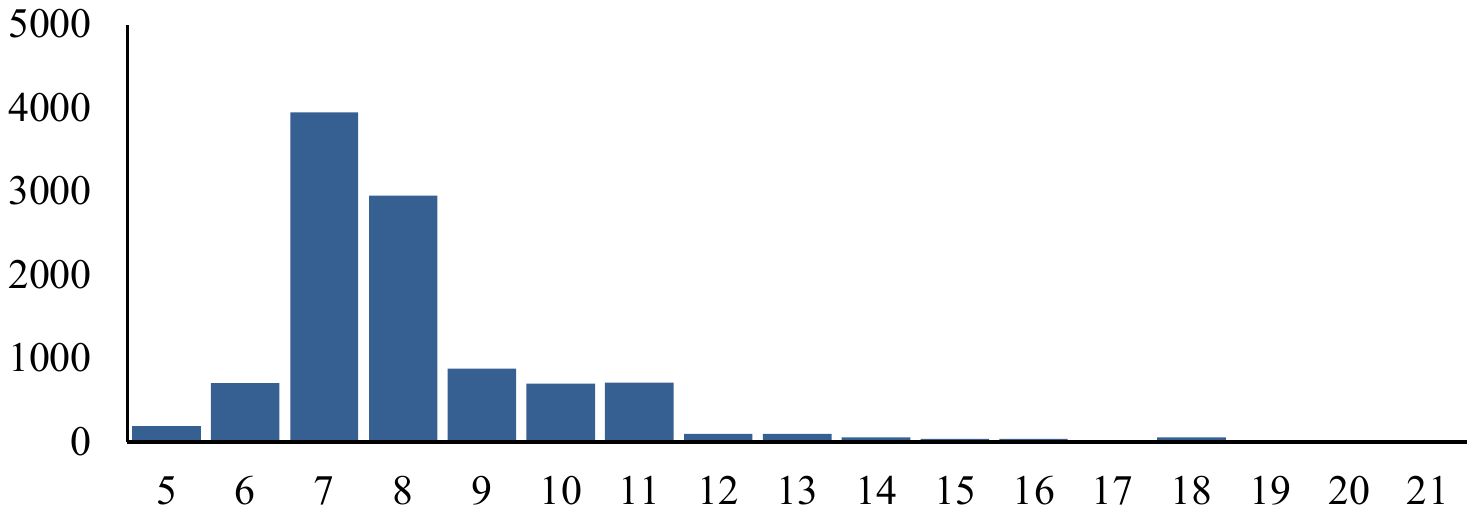}
    \vspace{-45pt}
    \caption{Length of breakpoint questions.}
    \label{fig:subfig-c}
  \end{subfigure}
  \caption{Length distribution of questions.}
  \label{fig:question}
\end{figure}

%% file: fig/answer.tex
\begin{figure}[t]
    \centering
    \begin{subfigure}[b]{0.4\textwidth}
    \includegraphics[width=1\linewidth]{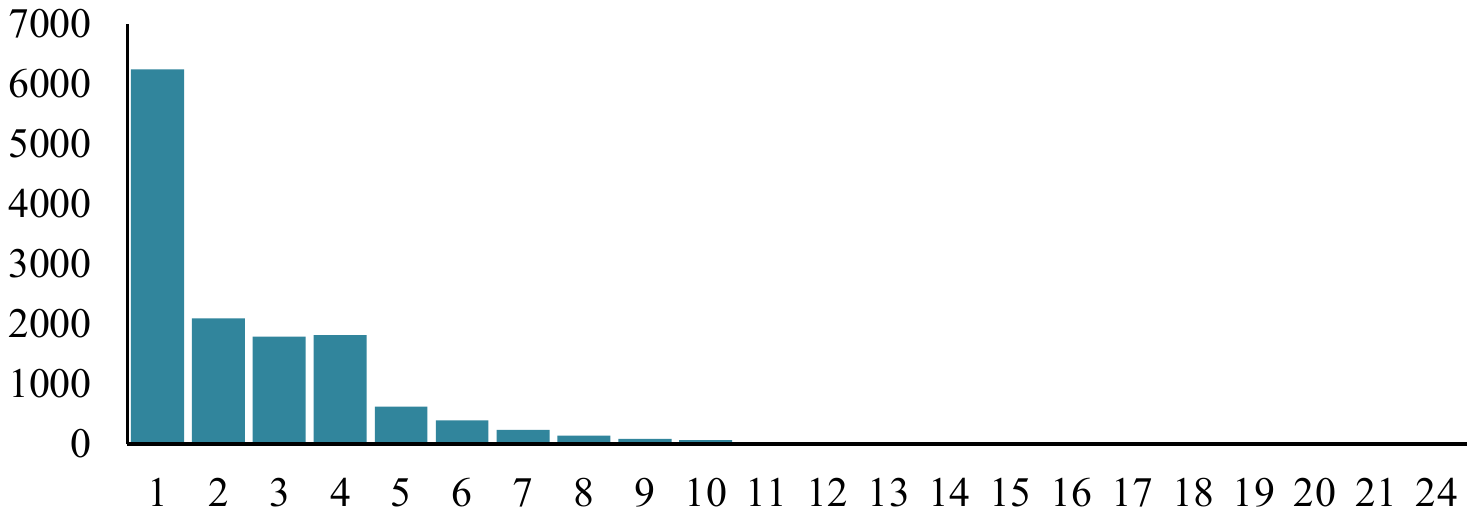}
    \vspace{-45pt}
    \caption{Length of total answers.}
    \label{fig:subfig-a}
  \end{subfigure}
  \begin{subfigure}[b]{0.4\textwidth}
    \vspace{-30pt}
    \includegraphics[width=1\linewidth]{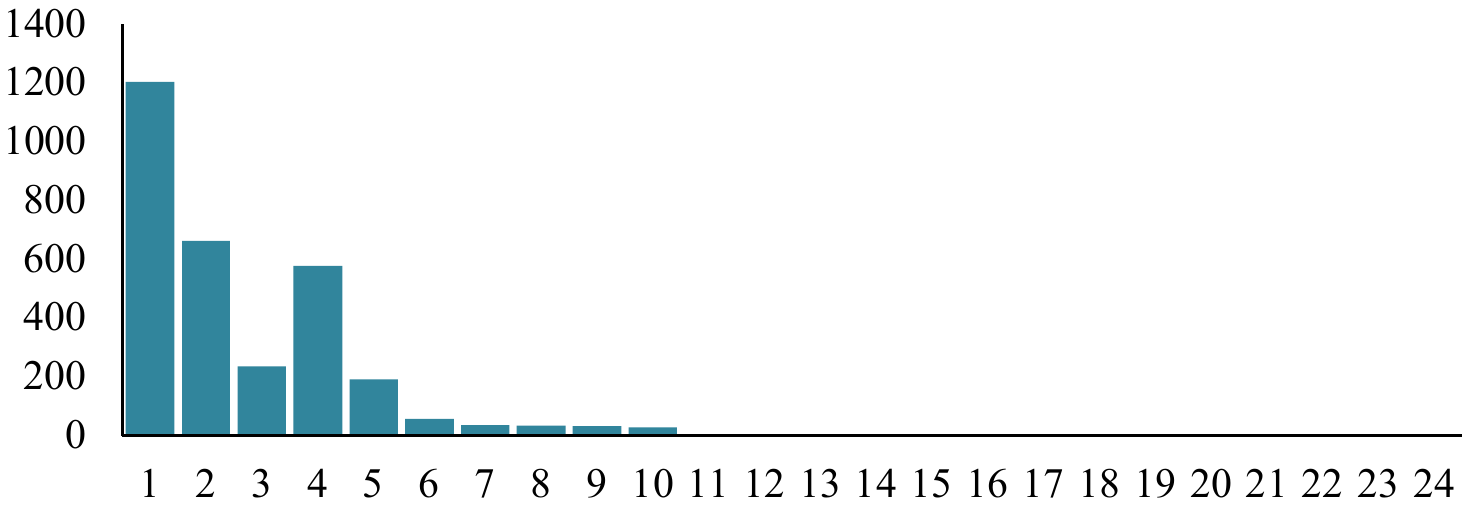}
    \vspace{-45pt}
    \caption{Length of global answers.}
    \label{fig:subfig-b}
  \end{subfigure}
  \begin{subfigure}[b]{0.4\textwidth}
    \vspace{-30pt}
    \includegraphics[width=1\linewidth]{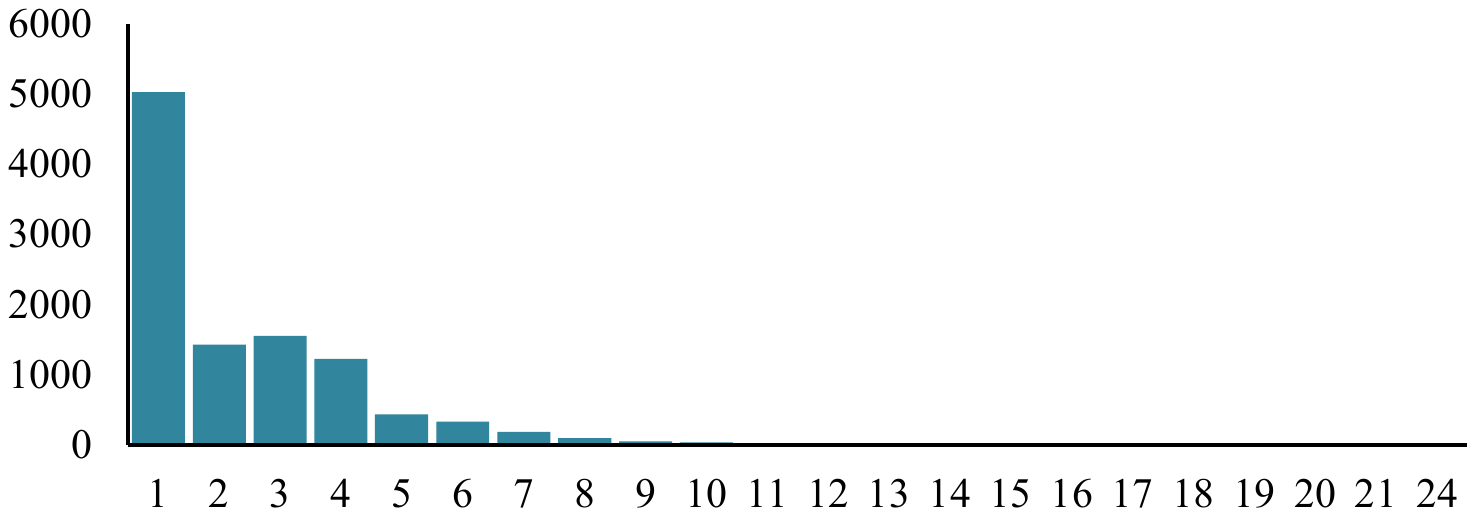}
    \vspace{-45pt}
    \caption{Length of breakpoint answers.}
    \label{fig:subfig-c}
  \end{subfigure}
  \caption{Length distribution of answers.}
  \label{fig:answer}
\end{figure}

%% file: fig/caption_length.tex
\begin{figure}[t]
    \centering
    \includegraphics[width=0.6\linewidth]{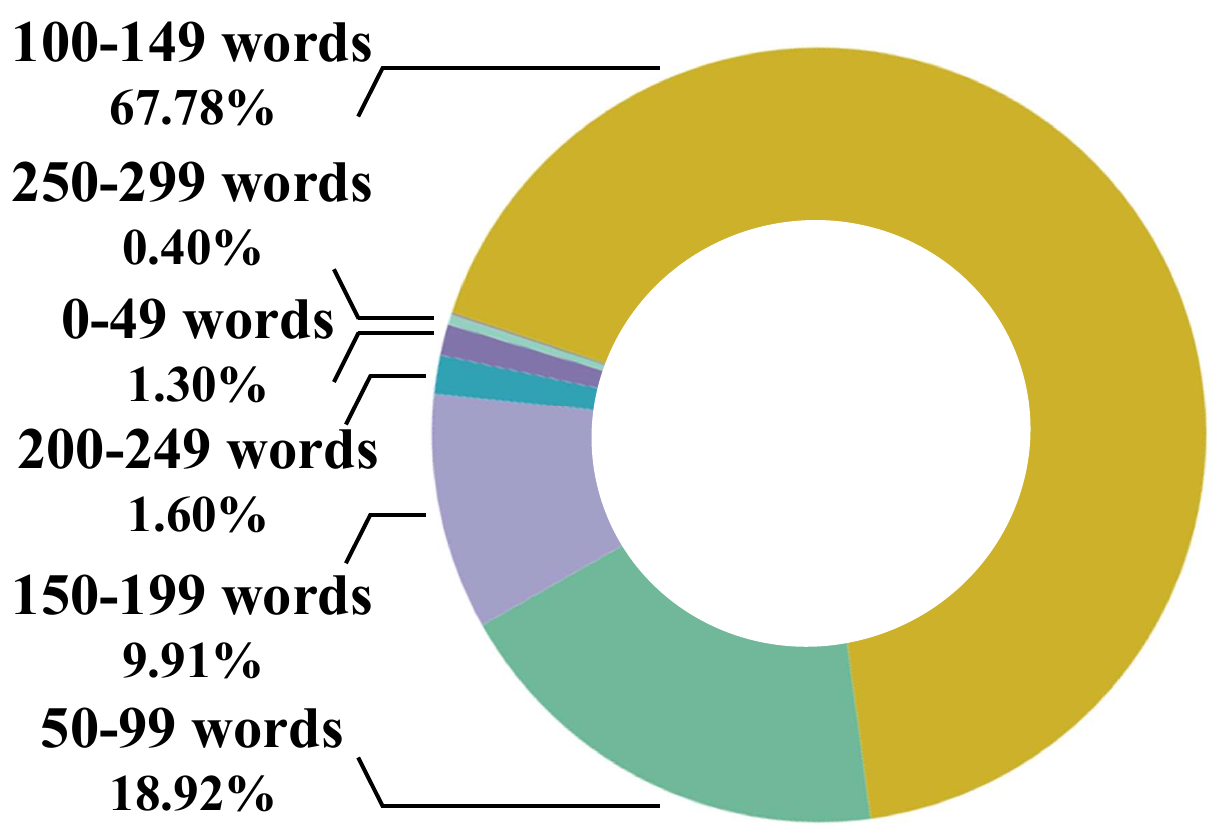}
    \caption{\textbf{Distribution of caption length.}}
    \label{fig:caption_length}
\end{figure}

%% file: fig/caption_wordcloud.tex
\begin{figure}[h]
    \centering
    \includegraphics[width=1\linewidth]{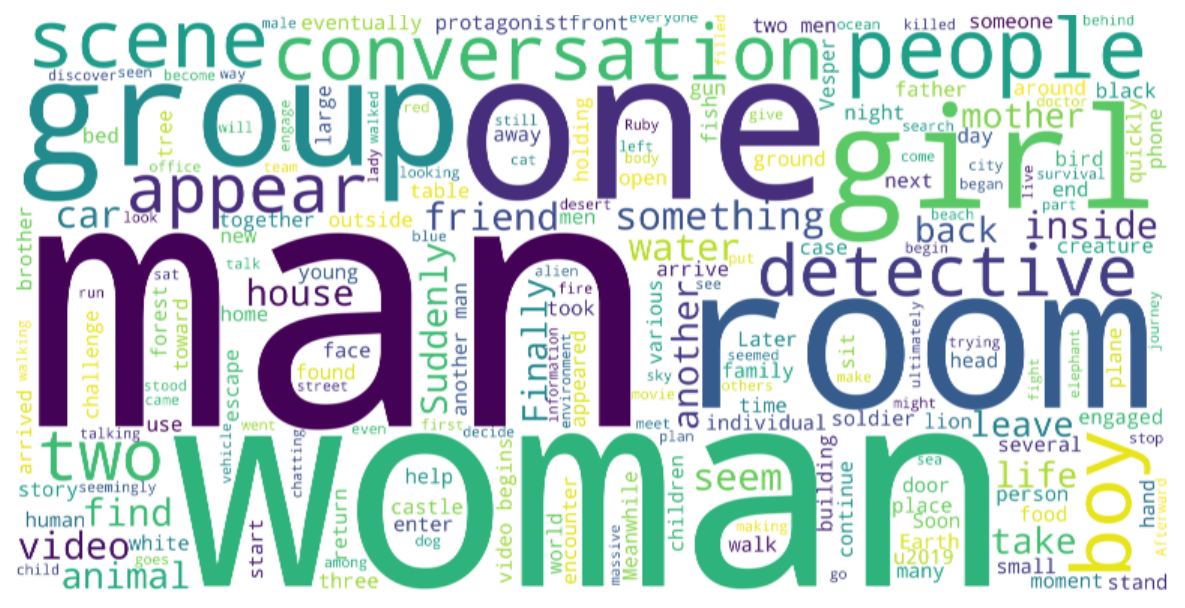}
    \caption{\textbf{Word Cloud} of the caption set in MovieChat-1K.}
    \label{fig:caption_wordcloud}
\end{figure}

%% file: tab/benchmarks.tex
\begin{table*}

\centering
\Large
\setlength{\tabcolsep}{8pt}
\renewcommand{\arraystretch}{1.3}
\resizebox{0.95\linewidth}{!}{
\begin{tabular}{ c | c | c | c | c | c | c}
\toprule
Dataset & Avg. Duration (min) & Number of Captions & Avg. Caption Length & Number of Question-Answer Pairs & Avg. Question Length& Avg. Answer Length \\

\midrule
MovieQA~\cite{tapaswi2016movieqa} & 3.5 & - & - & 14.9K& 9.3& 5.1\\
MovieGraphs~\cite{vicol2018moviegraphs} & 0.73 & 15K& 35& - & - & -\\
MovieNet~\cite{huang2020movienet} & 2.1 & 2.5K & - & - & - & - \\
\midrule
MovieChat-1K & 9.4 & 1K & 121 & 13K& 7.8& 2.3\\
\bottomrule
\end{tabular}
}

\caption{Comparison between MovieChat-1K and other benchmarks.}
\label{tab:benchmarks}    
\end{table*}

%% file: fig/prompt.tex
\begin{figure*}[h]
    \centering
    \includegraphics[width=1\linewidth]{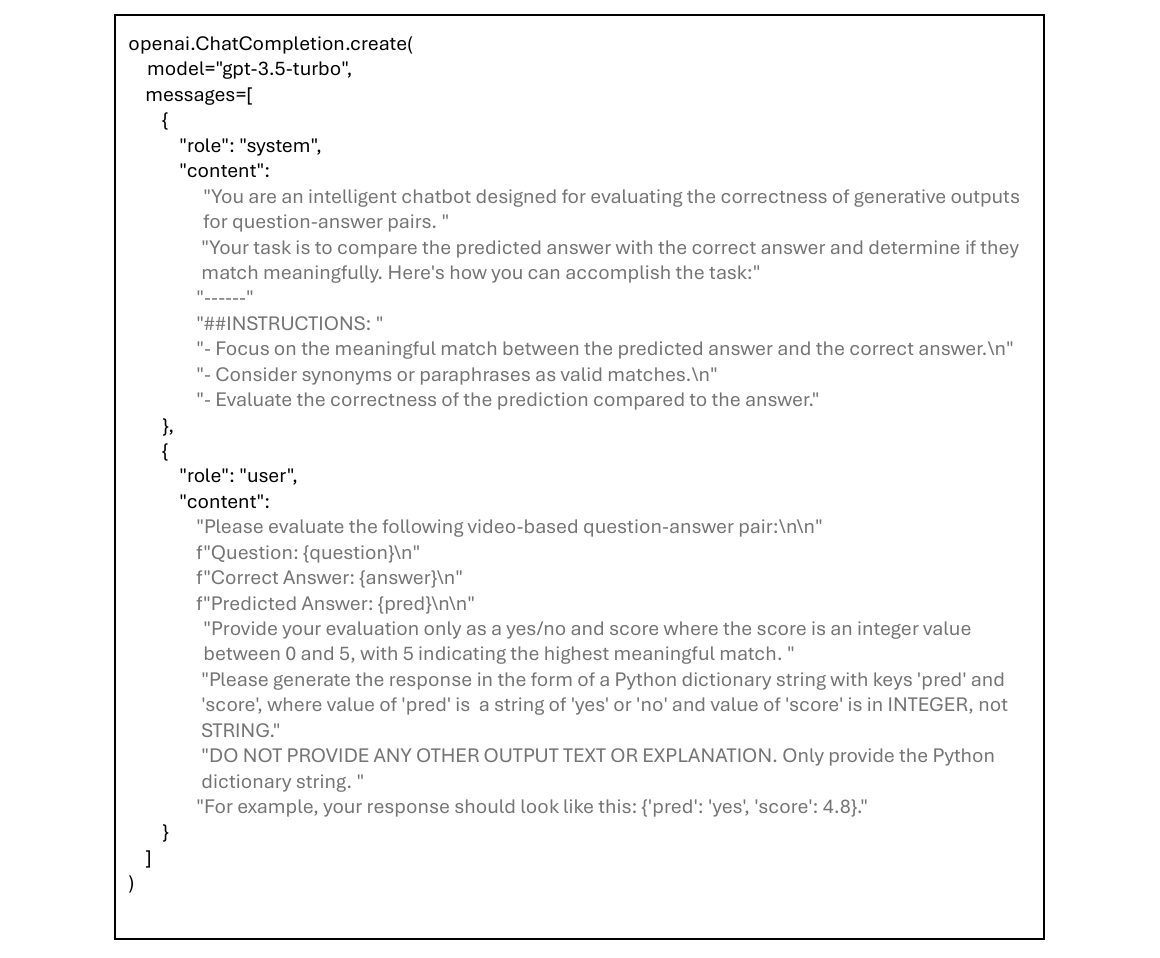}
    \caption{Prompt for ChatGPT in LLM-Assisted Evaluation for the short video question-answering task.}
    \label{fig:prompt}
\end{figure*}

%% file: tab/hyper-param.tex
\begin{table}[h]
\centering
\Large
\setlength{\tabcolsep}{8pt}
\renewcommand{\arraystretch}{1.3}
\resizebox{0.95\linewidth}{!}{
\begin{tabular}{ c | c}
\toprule
Description & Default Value \\
\midrule
size of sliding window & $16$ frames\\
size of short-term memory & $18$ frames $\times$ $32$ tokens per frames\\
size of long-term memory & $256$ frames\\
consolidation length & 2 \\
\bottomrule
\end{tabular}
}
\caption{Hyper-parameter settings of MovieChat.}
\label{tab:hyper}
\end{table}

%% file: tab/llm_score.tex
\begin{table}[h]
\centering
\setlength{\tabcolsep}{9pt}
\renewcommand{\arraystretch}{0.7}
\resizebox{\linewidth}{!}{
\begin{tabular}{@{} l c c c c @{}}
\toprule
\multirow{2}{*}{\vspace{-0.5ex} \tiny \textbf{Method}} & \multicolumn{2}{c}{\tiny \textbf{Global Mode}} & \multicolumn{2}{c}{\tiny \textbf{Breakpoint Mode}} \\
\cline{2-5}
 & \tiny \textbf{Accuracy} & \tiny \textbf{Score} & \tiny \textbf{Accuracy} & \tiny \textbf{Score}\\
\midrule
\vspace{-0.5ex} \tiny LLama~\cite{touvron2023llama} & \tiny \textbf{67.8}& \tiny \textbf{3.81}& \tiny \textbf{50.4}& \tiny 2.96 \\
\vspace{-0.5ex} \tiny LLama2~\cite{touvron2023llama2}& \tiny 64.2& \tiny 3.79& \tiny 48.1& \tiny \textbf{2.98} \\
\bottomrule
\end{tabular}
}
\caption{Ablation Study on how LLM affects the long video question answering. The best result is highlighted in bold.}
\label{tab:llm_score}
\end{table}

%% file: tab/llm_5.tex
\begin{table}[h]
\centering
\setlength{\tabcolsep}{6pt}
\renewcommand{\arraystretch}{1.3}
\resizebox{\linewidth}{!}{
\begin{tabular}{@{} l c c c c c c c c c c @{}}
\toprule
\multirow{2}{*}{\textbf{Method}} & \multicolumn{5}{c}{\textbf{Global Mode}} & \multicolumn{5}{c}{\textbf{Breakpoint Mode}} \\
\cline{2-11}
& \textbf{CI} & \textbf{DO} & \textbf{CU} & \textbf{TU} & \textbf{CO} & \textbf{CI} & \textbf{DO} & \textbf{CU} & \textbf{TU} & \textbf{CO}\\
\midrule
\midrule
LLama~\cite{touvron2023llama}&  \textbf{3.32}&  \textbf{3.28}&  3.40&  \textbf{2.97}& \textbf{3.48}& \textbf{2.97}& \textbf{3.24}& 3.31& \textbf{2.70}& 3.45 \\
LLama2~\cite{touvron2023llama2}&  3.27&  \textbf{3.28}&  \textbf{3.41} &  2.95& 3.45 & 2.96& 3.12& \textbf{3.38}& 2.68& \textbf{3.34}\\
\bottomrule
\end{tabular}
}
\caption{Ablation Study on how the large language model affects the long video generative performance. MM stands for memory mechanism, CI stands for correctness of information, DO stands for detail orientation, CU stands for contextual understanding, TU stands for temporal understanding, and CO stands for consistency.The best result is highlighted in bold.}
\label{tab:llm_5}
\end{table}

%% file: fig/llm.tex
\definecolor{global}{RGB}{21,96,130}
\definecolor{breakpoint}{RGB}{51,0,111}
\definecolor{hall}{RGB}{183,165,122}

\begin{figure*}[!t]
	\centering
    \includegraphics[width=0.95\linewidth]{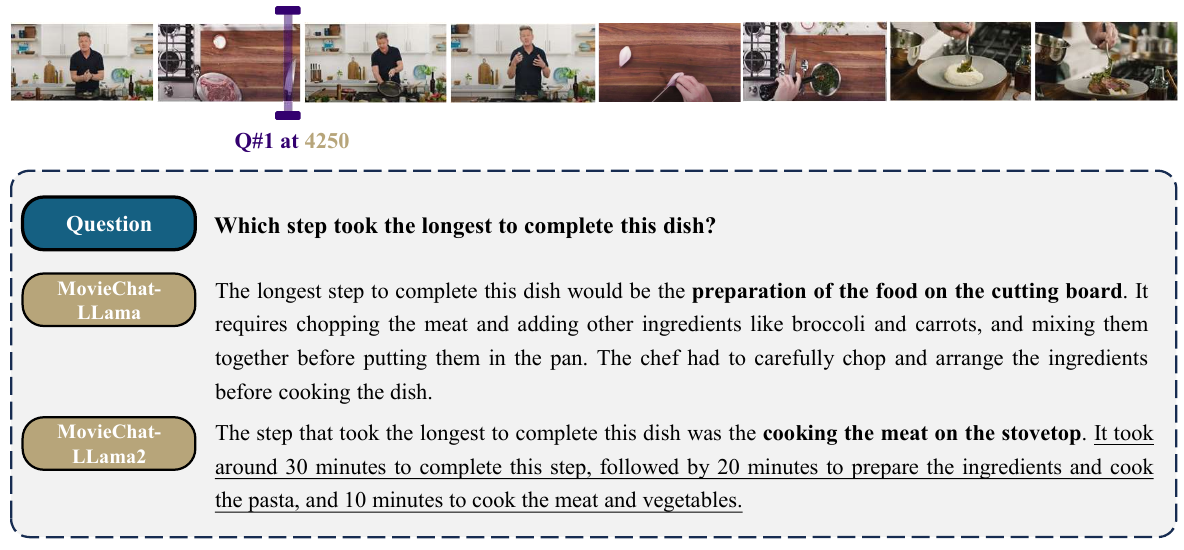}
	\caption{Question and answer about clips from \textit{YouTube}, which is a tutorial on how to cook steak. When asked ``Which step took the longest to complete this dish?" in global mode, MovieChat powered by different large language models provided divergent answers.}
 \vspace{20pt}
\label{fig:llm}
\end{figure*}

%% file: tab/type_result_global.tex
\begin{table}[t]
\centering
\setlength{\tabcolsep}{9pt}
\renewcommand{\arraystretch}{1.3}
\resizebox{\linewidth}{!}{
\begin{tabular}{@{} l c c c c c c @{}}
\toprule
\multirow{2}{*}{\vspace{-1ex} \textbf{Method}} & \multicolumn{2}{c}{ \textbf{Total}} & \multicolumn{2}{c}{ \textbf{Multi-choice}} & \multicolumn{2}{c}{ \textbf{Open-ended}} \\
\cline{2-7}
 &  \textbf{Accuracy} &  \textbf{Score} &  \textbf{Accuracy} &  \textbf{Score} &  \textbf{Accuracy} &  \textbf{Score}\\
\midrule
Video Chat~\cite{li2023videochat}& \underline{61.0}& \underline{3.34} & 74.8 & \underline{3.83} & \underline{56.4} & \underline{3.02} \\
Video LLaMA~\cite{zhang2023video}& 51.4&3.10 & \underline{78.3} & 3.58 & 38.8 & 2.67 \\
Video-ChatGPT~\cite{maaz2023video}& 44.2& 2.71 & 52.5 & 3.16 & 37.7 & 2.54 \\ 
\midrule
MovieChat~\textit{(ours)} & \textbf{67.8} & \textbf{3.81} &\textbf{ 80.9} & \textbf{4.02} & \textbf{57.5} & \textbf{3.74}\\
\bottomrule
\end{tabular}
}
\caption{Quantitative evaluation for long video different types question answering in global mode. The best result is highlighted in bold, and the second best is underlined.}
\label{tab:type_result_global}
\end{table}

%% file: tab/type_result_break.tex
\begin{table}[t]
\centering
\setlength{\tabcolsep}{9pt}
\renewcommand{\arraystretch}{1.3}
\resizebox{\linewidth}{!}{
\begin{tabular}{@{} l c c c c c c @{}}
\toprule
\multirow{2}{*}{\vspace{-1ex} \textbf{Method}} & \multicolumn{2}{c}{ \textbf{Total}} & \multicolumn{2}{c}{ \textbf{Multi-choice}} & \multicolumn{2}{c}{ \textbf{Open-ended}} \\
\cline{2-7}
 &  \textbf{Accuracy} &  \textbf{Score} &  \textbf{Accuracy} &  \textbf{Score} &  \textbf{Accuracy} &  \textbf{Score}\\
\midrule
Video Chat~\cite{li2023videochat}& 48.3& 2.43 & \textbf{62.4} & \underline{3.46} & 44.5 & 2.19 \\
Video LLaMA~\cite{zhang2023video}& 38.2& 2.33 & 57.3 & 2.39 & 33.1 & 2.31 \\
Video-ChatGPT~\cite{maaz2023video}& \underline{49.8}& \underline{2.71} & \underline{58.3} & 3.05 & \underline{47.5} & \underline{2.37} \\ 
\midrule
MovieChat~\textit{(ours)} & \textbf{50.2} & \textbf{2.96} & \textbf{62.4} & \textbf{3.65} & \textbf{46.7} & \textbf{2.70}\\
\bottomrule
\end{tabular}
}
\caption{Quantitative evaluation for long video different types question answering in breakpoint mode. The best result is highlighted in bold, and the second best is underlined.}
\label{tab:type_result_break}
\end{table}

%% file: tab/break_5.tex
\begin{table}[t]
\centering
\setlength{\tabcolsep}{9pt}
\renewcommand{\arraystretch}{0.8}
\resizebox{\linewidth}{!}{
\begin{tabular}{@{} l c c c c c c @{}}
\toprule
\scriptsize \textbf{Method} & \scriptsize \textbf{CI} & \scriptsize \textbf{DO} & \scriptsize \textbf{CU} & \scriptsize \textbf{TU} & \scriptsize \textbf{CO}\\
\midrule
\scriptsize Video Chat~\cite{li2023videochat} & \scriptsize 2.42 & \scriptsize 2.51 & \scriptsize 2.81 & \scriptsize 2.10 & \scriptsize 2.78\\
\scriptsize Video LLaMA~\cite{zhang2023video} & \scriptsize 2.04 & \scriptsize 2.29& \scriptsize 2.63& \scriptsize 2.00& \scriptsize 2.87\\
\scriptsize Video-ChatGPT~\cite{maaz2023video}& \scriptsize \underline{2.62}& \scriptsize \textbf{2.65}& \scriptsize \underline{2.86}& \scriptsize \underline{2.32}& \scriptsize \underline{2.96}\\
\midrule
\scriptsize MovieChat~\textit{(Ours)}& \scriptsize \textbf{2.64}& \scriptsize \underline{2.60}& \scriptsize \textbf{2.87}& \scriptsize \textbf{2.49}& \scriptsize \textbf{3.08} \\
\bottomrule
\end{tabular}
}
\caption{Quantitative evaluation for long video generation performance in breakpoint mode with the average of GPT-3.5~\cite{gpt3.5}, Claude~\cite{examplewebpage} and human blind rating. CI stands for correctness of information, DO stands for detail orientation, CU stands for contextual understanding, TU stands for temporal understanding, and CO stands for consistency. The best result is highlighted in bold, and the second best is underlined.}
\label{tab:break_5}
\end{table}

%% file: fig/pear.tex
\begin{figure*}[t]
  \centering
  \begin{subfigure}[b]{0.32\textwidth}
    \centering
    \includegraphics[width=0.95\textwidth]{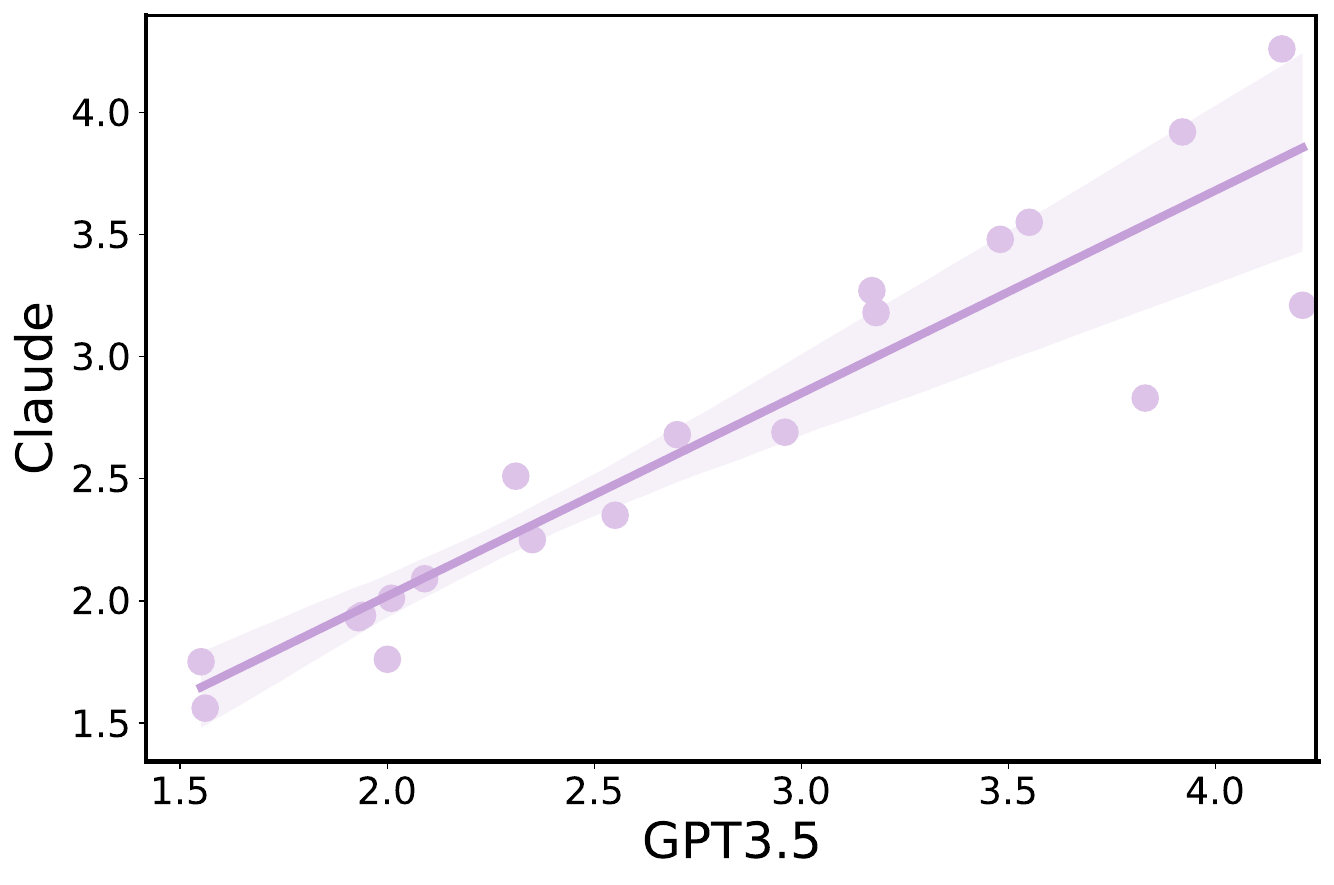}
    \caption{\textbf{GPT-3.5 VS. Claude} \\ \centering PCC = 0.927}
  \end{subfigure}
  \begin{subfigure}[b]{0.32\textwidth}
    \centering
    \includegraphics[width=0.95\textwidth]{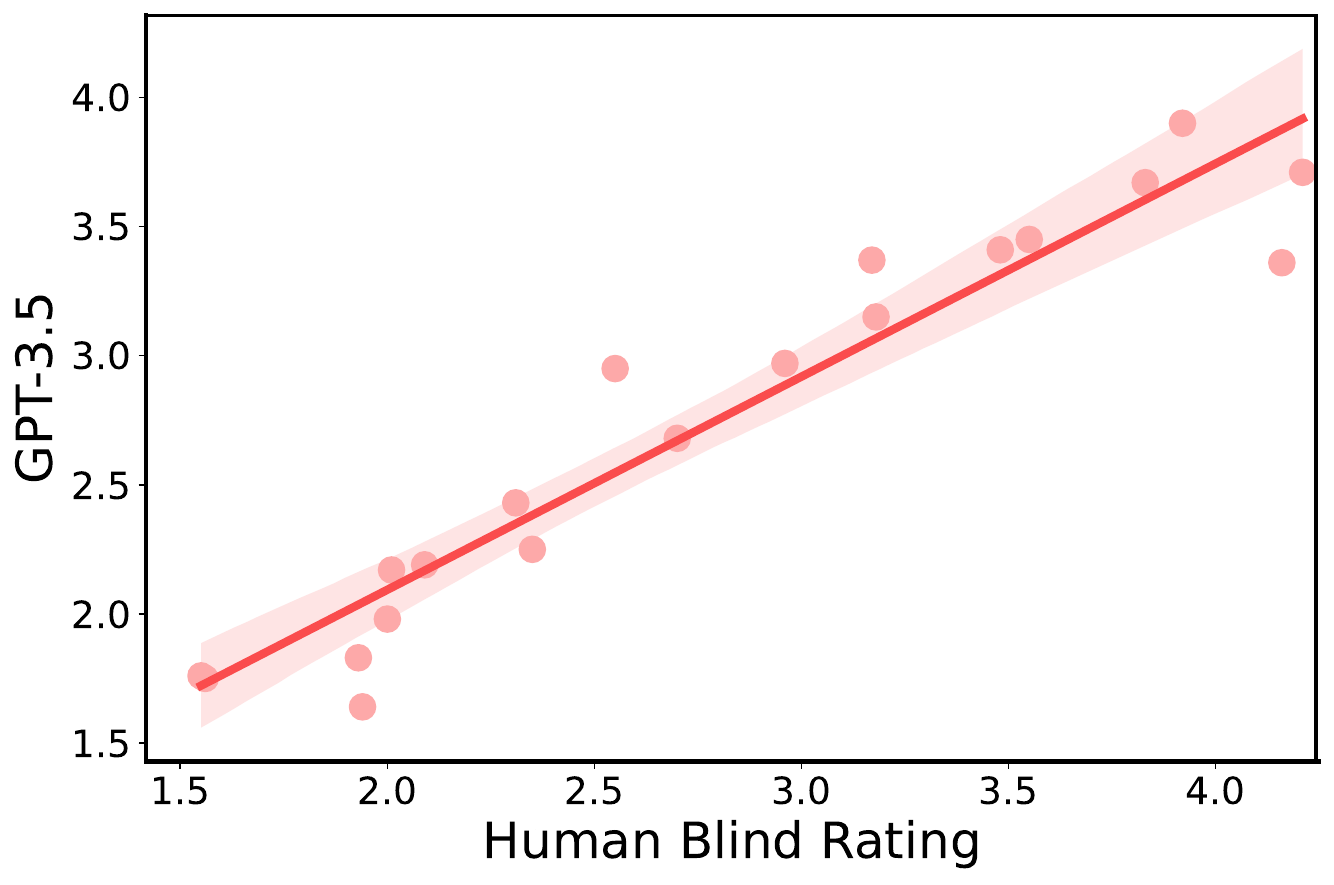}
    \caption{\textbf{GPT-3.5 VS. Human Blind Rating} \\ \centering PCC = 0.955}
  \end{subfigure}
  \begin{subfigure}[b]{0.32\textwidth}
    \centering
    \includegraphics[width=0.95\textwidth]{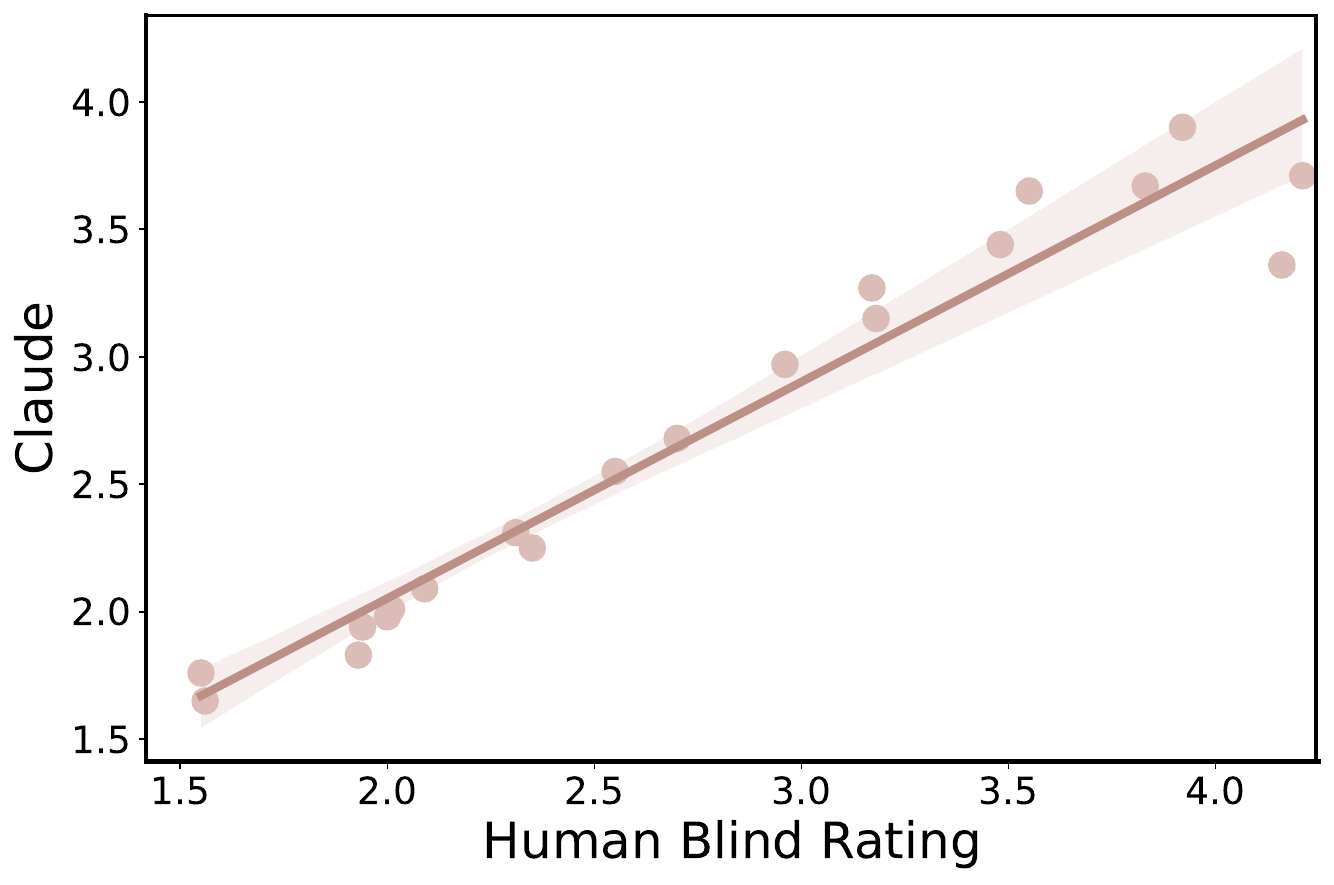}
    \caption{\textbf{Claude VS. Human Blind Rating} \\ \centering PCC = 0.978}
  \end{subfigure}
  \caption{Results of the Pearson correlation analysis between three evaluation methods, including GPT-3.5~\cite{gpt3.5}, Claude~\cite{examplewebpage}, and human blind rating. PCC stands for Pearson Correlation Coefficient.}
  \label{fig:pearson}
\end{figure*}

%% file: tab/Pearson.tex
\begin{table}[h]
\centering
\Large
\setlength{\tabcolsep}{8pt}
\renewcommand{\arraystretch}{1.3}
\resizebox{0.95\linewidth}{!}{
\begin{tabular}{ l | c }
\toprule
Evaluation Method & Pearson Correlation Coefficient\\
\midrule

GPT3.5 VS. Claude & 0.927\\
GPT3.5 VS. Human Blind Rating & 0.955\\
Claude VS. Human Blind Rating & 0.978\\
\bottomrule
\end{tabular}
}
\caption{Pearson correlation coefficient of GPT-3.5~\cite{gpt3.5}, Claude~\cite{examplewebpage}, and human blind rating on score. We calculate the mean score across each score dimensions for MovieChat and previous methods~\cite{maaz2023video,li2023videochat,zhang2023llama, zhang2023video}, and then computes the Pearson correlation between these means for each pair of evaluation methods. The Pearson correlation coefficient, which ranges from -1 to +1, indicates a stronger positive linear relationship between the two sets of data when the coefficient is higher (closer to +1).}
\label{tab:pearson}
\end{table}

%% file: tab/long_gpt.tex
\begin{table}[t]
\centering
\setlength{\tabcolsep}{8pt}
\renewcommand{\arraystretch}{1.3}
\resizebox{\linewidth}{!}{
\begin{tabular}{@{} l c c c c c @{}}
\toprule
\multirow{2}{*}{\textbf{Method}} & \multirow{2}{*}{\textbf{\# Frames}} & \multicolumn{2}{c}{\textbf{Global Mode}} & \multicolumn{2}{c}{\textbf{Breakpoint Mode}} \\
\cline{3-6}
 & & \textbf{Accuracy} & \textbf{Score} & \textbf{Accuracy} & \textbf{Score} \\
\midrule
Video Chat~\cite{li2023videochat}& 32 & \underline{61.0} & \underline{3.34} & 48.3 & 2.43 \\
Video LLaMA~\cite{zhang2023video}& 32 & 51.4 & 3.10 & 38.2 & 2.31 \\
Video-ChatGPT~\cite{maaz2023video}& 100 & 44.2 & 2.71 & \underline{49.8} & \underline{2.71} \\ 
\midrule
MovieChat~\textit{(ours)} & 2048 & \textbf{67.8} & \textbf{3.81} & \textbf{50.4} & \textbf{2.96}\\
\bottomrule
\end{tabular}
}
\caption{Quantitative evaluation for long video question answering on MovieChat-1K test set with GPT-3.5~\cite{gpt3.5}. The best result is highlighted in bold, and the second best is underlined.}
\label{tab:long_gpt}
\end{table}

%% file: tab/long_claude.tex
\begin{table}[t]
\centering
\setlength{\tabcolsep}{8pt}
\renewcommand{\arraystretch}{1.3}
\resizebox{\linewidth}{!}{
\begin{tabular}{@{} l c c c c c @{}}
\toprule
\multirow{2}{*}{\textbf{Method}} & \multirow{2}{*}{\textbf{\# Frames}} & \multicolumn{2}{c}{\textbf{Global Mode}} & \multicolumn{2}{c}{\textbf{Breakpoint Mode}} \\
\cline{3-6}
 & & \textbf{Accuracy} & \textbf{Score} & \textbf{Accuracy} & \textbf{Score} \\
\midrule
Video Chat~\cite{li2023videochat}& 32 & \underline{52.1} & \underline{2.59} &  \underline{43.8}& 2.12 \\
Video LLaMA~\cite{zhang2023video}& 32 &  47.3& 2.19 & 33.2 & 1.69 \\
Video-ChatGPT~\cite{maaz2023video}& 100 & 39.8 & 2.04 & \textbf{46.4} & \underline{2.21} \\ 
\midrule
MovieChat~\textit{(ours)} & 2048 & \textbf{55.3}&  \textbf{2.73}&  \textbf{46.4}& \textbf{2.28}\\
\bottomrule
\end{tabular}
}
\caption{Quantitative evaluation for long video question answering on MovieChat-1K test set with Claude~\cite{examplewebpage}. The best result is highlighted in bold, and the second best is underlined.}
\label{tab:long_claude}
\end{table}

%% file: tab/long_human.tex
\begin{table}[t]
\centering
\setlength{\tabcolsep}{8pt}
\renewcommand{\arraystretch}{1.3}
\resizebox{\linewidth}{!}{
\begin{tabular}{@{} l c c c c c @{}}
\toprule
\multirow{2}{*}{\textbf{Method}} & \multirow{2}{*}{\textbf{\# Frames}} & \multicolumn{2}{c}{\textbf{Global Mode}} & \multicolumn{2}{c}{\textbf{Breakpoint Mode}} \\
\cline{3-6}
 & & \textbf{Accuracy} & \textbf{Score} & \textbf{Accuracy} & \textbf{Score} \\
\midrule
Video Chat~\cite{li2023videochat}& 32 & \underline{60.2} & \underline{3.08} &  46.3& 2.32 \\
Video LLaMA~\cite{zhang2023video}& 32 &  56.3& 2.72 & 45.8 & 2.11 \\
Video-ChatGPT~\cite{maaz2023video}& 100 & 58.7 & 2.89 & \underline{47.8} & \underline{2.43} \\ 
\midrule
MovieChat~\textit{(ours)} & 2048 & \textbf{63.7}&  \textbf{3.15}&  \textbf{48.1}& \textbf{2.46}\\
\bottomrule
\end{tabular}
}
\caption{Quantitative evaluation for long video question answering on MovieChat-1K test set with human blind rating. The best result is highlighted in bold, and the second best is underlined.}
\label{tab:long_human}
\end{table}

%% file: tab/long_varies_gpt.tex
\begin{table}[t]
\centering
\setlength{\tabcolsep}{8pt}
\renewcommand{\arraystretch}{0.7}
\resizebox{\linewidth}{!}{
\begin{tabular}{@{} l c c c c c c @{}}
\toprule
\scriptsize \textbf{Method} & \scriptsize \textbf{CI} & \scriptsize \textbf{DO} & \scriptsize \textbf{CU} & \scriptsize \textbf{TU} & \scriptsize \textbf{CO}\\
\midrule
\scriptsize Video Chat~\cite{li2023videochat} & \scriptsize 3.26 & \scriptsize \underline{3.20} & \scriptsize \underline{3.38} & \scriptsize \underline{2.97}& \scriptsize  \underline{3.47}\\
\scriptsize Video LLaMA~\cite{zhang2023video} & \scriptsize \underline{3.30} & \scriptsize 2.53& \scriptsize 3.28& \scriptsize 2.77& \scriptsize 3.42\\
\scriptsize Video-ChatGPT~\cite{maaz2023video}& \scriptsize 2.48& \scriptsize 2.78& \scriptsize 3.03& \scriptsize 2.48& \scriptsize 2.99\\
\midrule
\scriptsize MovieChat~\textit{(Ours)}& \scriptsize \textbf{3.32}& \scriptsize \textbf{3.28} & \scriptsize \textbf{3.44}& \scriptsize \textbf{3.06}& \scriptsize \textbf{3.48} \\
\bottomrule
\end{tabular}
}
\caption{Quantitative evaluation for long video generation performance in global mode with GPT-3.5~\cite{gpt3.5}. CI stands for correctness of information, DO stands for detail orientation, CU stands for contextual understanding, TU stands for temporal understanding, and CO stands for consistency. The best result is highlighted in bold, and the second best is underlined.}
\label{tab:long_varies_gpt}
\end{table}

%% file: tab/long_varies_claude.tex
\begin{table}[t]
\centering
\setlength{\tabcolsep}{8pt}
\renewcommand{\arraystretch}{0.7}
\resizebox{\linewidth}{!}{
\begin{tabular}{@{} l c c c c c c @{}}
\toprule
\scriptsize \textbf{Method} & \scriptsize \textbf{CI} & \scriptsize \textbf{DO} & \scriptsize \textbf{CU} & \scriptsize \textbf{TU} & \scriptsize \textbf{CO}\\
\midrule
\scriptsize Video Chat~\cite{li2023videochat} & \scriptsize \underline{2.83} & \scriptsize \underline{2.43} & \scriptsize \underline{3.02} & \scriptsize \underline{2.87}& \scriptsize  \underline{2.93}\\
\scriptsize Video LLaMA~\cite{zhang2023video} & \scriptsize 2.04 & \scriptsize 1.66& \scriptsize 2.46& \scriptsize 2.07& \scriptsize 2.36\\
\scriptsize Video-ChatGPT~\cite{maaz2023video}& \scriptsize 1.81& \scriptsize 1.65& \scriptsize 2.05& \scriptsize2.07& \scriptsize 2.07\\
\midrule
\scriptsize MovieChat~\textit{(Ours)}& \scriptsize \textbf{2.88}& \scriptsize \textbf{2.82} & \scriptsize \textbf{3.11}& \scriptsize \textbf{3.04}& \scriptsize \textbf{2.96} \\
\bottomrule
\end{tabular}
}
\caption{Quantitative evaluation for long video generation performance in global mode with Claude~\cite{examplewebpage}. CI stands for correctness of information, DO stands for detail orientation, CU stands for contextual understanding, TU stands for temporal understanding, and CO stands for consistency. The best result is highlighted in bold, and the second best is underlined.}
\label{tab:long_varies_claude}
\end{table}

%% file: tab/long_varies_human.tex
\begin{table}[t]
\centering
\setlength{\tabcolsep}{8pt}
\renewcommand{\arraystretch}{0.7}
\resizebox{\linewidth}{!}{
\begin{tabular}{@{} l c c c c c c @{}}
\toprule
\scriptsize \textbf{Method} & \scriptsize \textbf{CI} & \scriptsize \textbf{DO} & \scriptsize \textbf{CU} & \scriptsize \textbf{TU} & \scriptsize \textbf{CO}\\
\midrule
\scriptsize Video Chat~\cite{li2023videochat} & \scriptsize \underline{3.03} & \scriptsize \underline{2.61} & \scriptsize \underline{2.87} & \scriptsize \underline{3.15}& \scriptsize  \underline{3.23}\\
\scriptsize Video LLaMA~\cite{zhang2023video} & \scriptsize 2.91 & \scriptsize 2.54& \scriptsize 2.74& \scriptsize 3.01& \scriptsize 3.12\\
\scriptsize Video-ChatGPT~\cite{maaz2023video}& \scriptsize 2.83& \scriptsize 2.47& \scriptsize 2.66& \scriptsize2.92& \scriptsize 3.01\\
\midrule
\scriptsize MovieChat~\textit{(Ours)}& \scriptsize \textbf{3.12}& \scriptsize \textbf{2.68} & \scriptsize \textbf{3.17}& \scriptsize \textbf{3.41}& \scriptsize \textbf{3.31} \\
\bottomrule
\end{tabular}
}
\caption{Quantitative evaluation for long video generation performance in global mode with human blind rating. CI stands for correctness of information, DO stands for detail orientation, CU stands for contextual understanding, TU stands for temporal understanding, and CO stands for consistency. The best result is highlighted in bold, and the second best is underlined.}
\label{tab:long_varies_claude}
\end{table}

%% file: tab/break_5_gpt.tex
\begin{table}[t]
\centering
\setlength{\tabcolsep}{9pt}
\renewcommand{\arraystretch}{0.8}
\resizebox{\linewidth}{!}{
\begin{tabular}{@{} l c c c c c c @{}}
\toprule
\scriptsize \textbf{Method} & \scriptsize \textbf{CI} & \scriptsize \textbf{DO} & \scriptsize \textbf{CU} & \scriptsize \textbf{TU} & \scriptsize \textbf{CO}\\
\midrule
\scriptsize Video Chat~\cite{li2023videochat} & \scriptsize 2.96 & \scriptsize 3.09 & \scriptsize 3.24& \scriptsize 2.46 & \scriptsize 3.22\\
\scriptsize Video LLaMA~\cite{zhang2023video} & \scriptsize 2.42 & \scriptsize 2.85& \scriptsize 2.87& \scriptsize 2.00& \scriptsize 2.87\\
\scriptsize Video-ChatGPT~\cite{maaz2023video}& \scriptsize \textbf{3.11}& \scriptsize \textbf{3.32}& \scriptsize \underline{3.29}& \scriptsize \underline{2.62}& \scriptsize \underline{3.29}\\
\midrule
\scriptsize MovieChat~\textit{(Ours)}& \scriptsize \underline{3.07}& \scriptsize \underline{3.24}& \scriptsize \textbf{3.31}& \scriptsize \textbf{2.70}& \scriptsize \textbf{3.45} \\
\bottomrule
\end{tabular}
}
\caption{Quantitative evaluation for long video generation performance in breakpoint mode with GPT-3.5~\cite{gpt3.5}. CI stands for correctness of information, DO stands for detail orientation, CU stands for contextual understanding, TU stands for temporal understanding, and CO stands for consistency. The best result is highlighted in bold, and the second best is underlined.}
\label{tab:break_5_gpt}
\end{table}

%% file: tab/break_5_claude.tex
\begin{table}[t]
\centering
\setlength{\tabcolsep}{9pt}
\renewcommand{\arraystretch}{0.8}
\resizebox{\linewidth}{!}{
\begin{tabular}{@{} l c c c c c c @{}}
\toprule
\scriptsize \textbf{Method} & \scriptsize \textbf{CI} & \scriptsize \textbf{DO} & \scriptsize \textbf{CU} & \scriptsize \textbf{TU} & \scriptsize \textbf{CO}\\
\midrule
\scriptsize Video Chat~\cite{li2023videochat} & \scriptsize 2.12 & \scriptsize 2.20 & \scriptsize 2.30& \scriptsize 1.97 & \scriptsize 2.37\\
\scriptsize Video LLaMA~\cite{zhang2023video} & \scriptsize 1.62 & \scriptsize 1.85& \scriptsize 2.20& \scriptsize 1.34& \scriptsize 2.02\\
\scriptsize Video-ChatGPT~\cite{maaz2023video}& \scriptsize \underline{2.36}& \scriptsize \textbf{2.26}& \scriptsize \underline{2.34}& \scriptsize \underline{2.23}& \scriptsize \textbf{2.70}\\
\midrule
\scriptsize MovieChat~\textit{(Ours)}& \scriptsize \textbf{2.38} & \scriptsize \underline{2.16}& \scriptsize \textbf{2.35}& \scriptsize \textbf{2.43}& \scriptsize \underline{2.68} \\
\bottomrule
\end{tabular}
}
\caption{Quantitative evaluation for long video generation performance in breakpoint mode with Claude~\cite{examplewebpage}. CI stands for correctness of information, DO stands for detail orientation, CU stands for contextual understanding, TU stands for temporal understanding, and CO stands for consistency. The best result is highlighted in bold, and the second best is underlined.}
\label{tab:break_5_claude}
\end{table}

%% file: tab/break_5_human.tex
\begin{table}[t]
\centering
\setlength{\tabcolsep}{9pt}
\renewcommand{\arraystretch}{0.8}
\resizebox{\linewidth}{!}{
\begin{tabular}{@{} l c c c c c c @{}}
\toprule
\scriptsize \textbf{Method} & \scriptsize \textbf{CI} & \scriptsize \textbf{DO} & \scriptsize \textbf{CU} & \scriptsize \textbf{TU} & \scriptsize \textbf{CO}\\
\midrule
\scriptsize Video Chat~\cite{li2023videochat} & \scriptsize 2.17 & \scriptsize 2.24 & \scriptsize 2.89& \scriptsize 1.87 & \scriptsize 2.75\\
\scriptsize Video LLaMA~\cite{zhang2023video} & \scriptsize 2.09 & \scriptsize 2.18& \scriptsize 2.82 & \scriptsize 1.74& \scriptsize 2.68\\
\scriptsize Video-ChatGPT~\cite{maaz2023video}& \scriptsize \underline{2.39}& \scriptsize \underline{2.36}& \scriptsize \textbf{2.96}& \scriptsize \underline{2.10}& \scriptsize \underline{2.89}\\
\midrule
\scriptsize MovieChat~\textit{(Ours)}& \scriptsize \textbf{2.48}& \scriptsize \textbf{2.41}& \scriptsize \underline{2.94}& \scriptsize \textbf{2.33}& \scriptsize \textbf{3.12} \\
\bottomrule
\end{tabular}
}
\caption{Quantitative evaluation for long video generation performance in breakpoint mode with human blind rating. CI stands for correctness of information, DO stands for detail orientation, CU stands for contextual understanding, TU stands for temporal understanding, and CO stands for consistency. The best result is highlighted in bold, and the second best is underlined.}
\label{tab:break_5_human}
\end{table}

%% file: fig/case1.tex
\begin{figure*}[!t]
	\centering
    \includegraphics[width=0.95\linewidth]{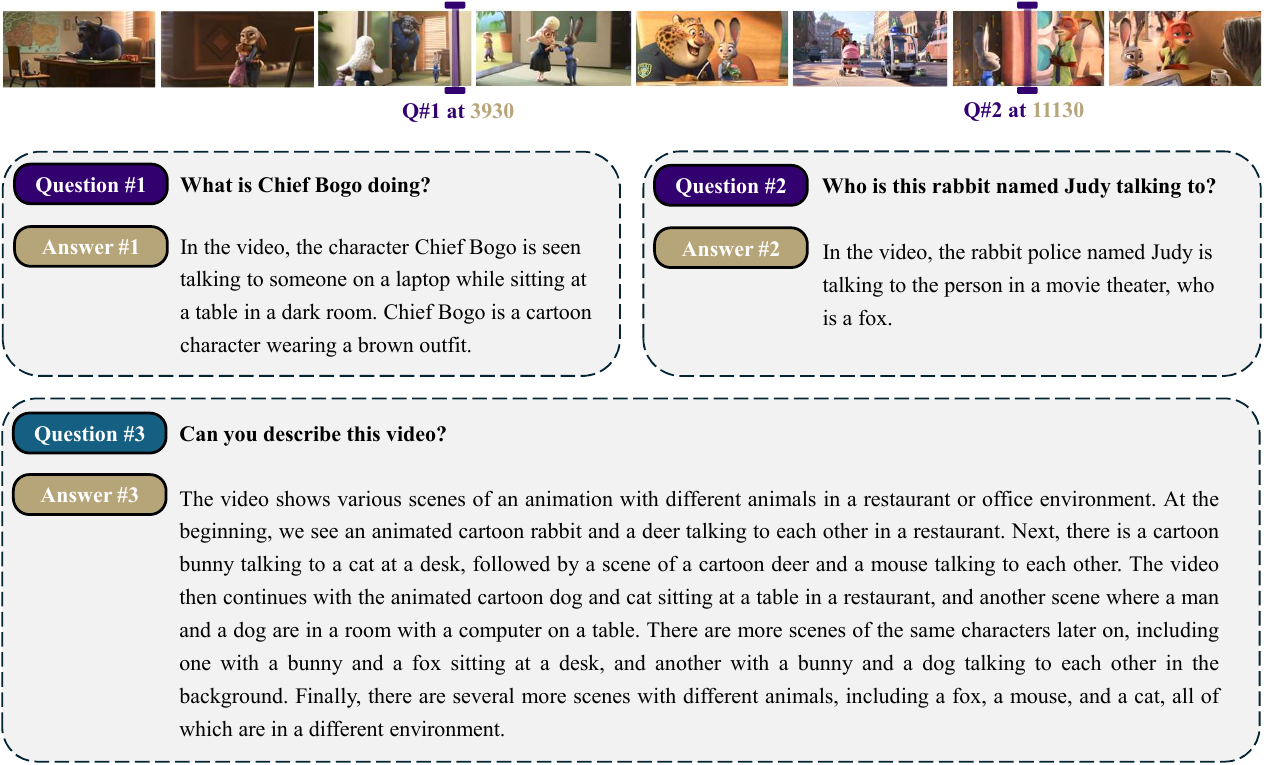}
	\caption{Question and answer about clips from \textit{Zootopia}, a cartoon, which tells the story of a determined police officer rabbit named Judy who pairs up with a cunning fox to uncover a conspiracy about missing animals and develop an unexpected friendship.}
        \label{fig:case1}
    \vspace{20pt}
    \includegraphics[width=0.95\linewidth]{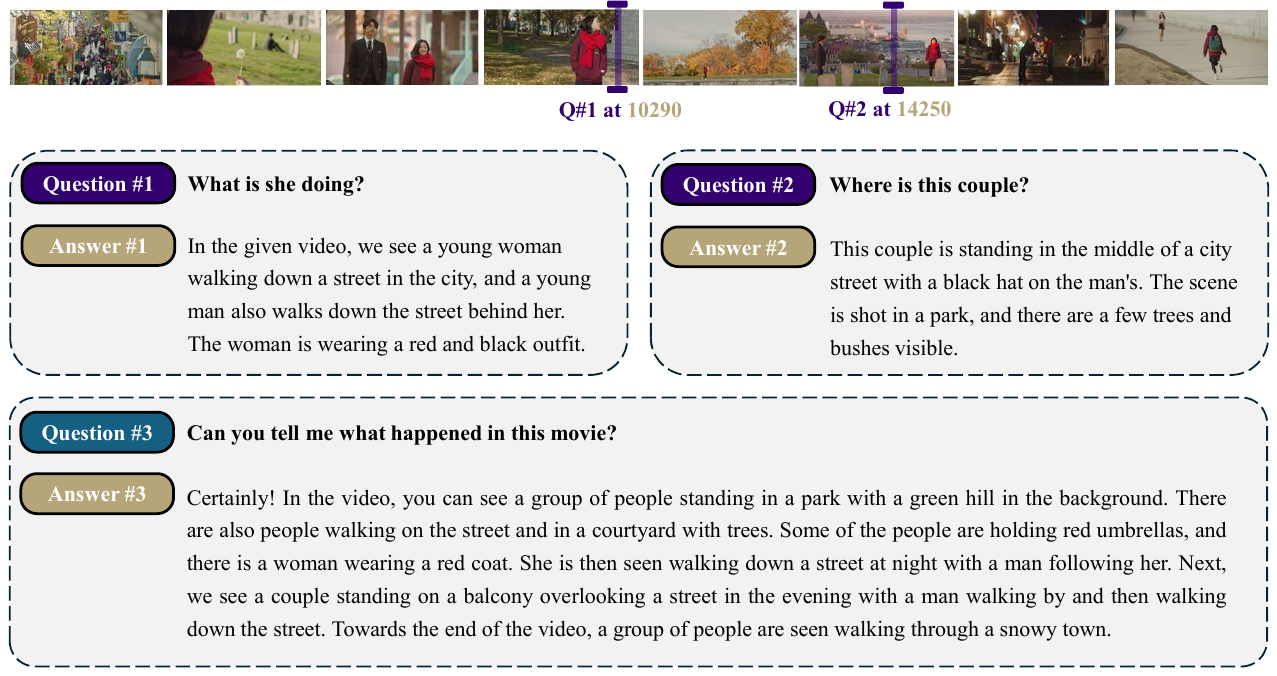}
	\caption{Question and answer about clips from \textit{Goblin}, which tells the story of Kim Shin, an immortal "goblin" who needs to find a human bride to end his endless life but instead meets Ji Eun-tak, a girl fated to die who claims to be the "goblin's bride," leading to a romantic tale unfolding between them.}
        \label{fig:case2}
 \vspace{20pt}

\end{figure*}

%% file: fig/case2.tex
\begin{figure*}[!t]
	\centering
    \includegraphics[width=0.95\linewidth]{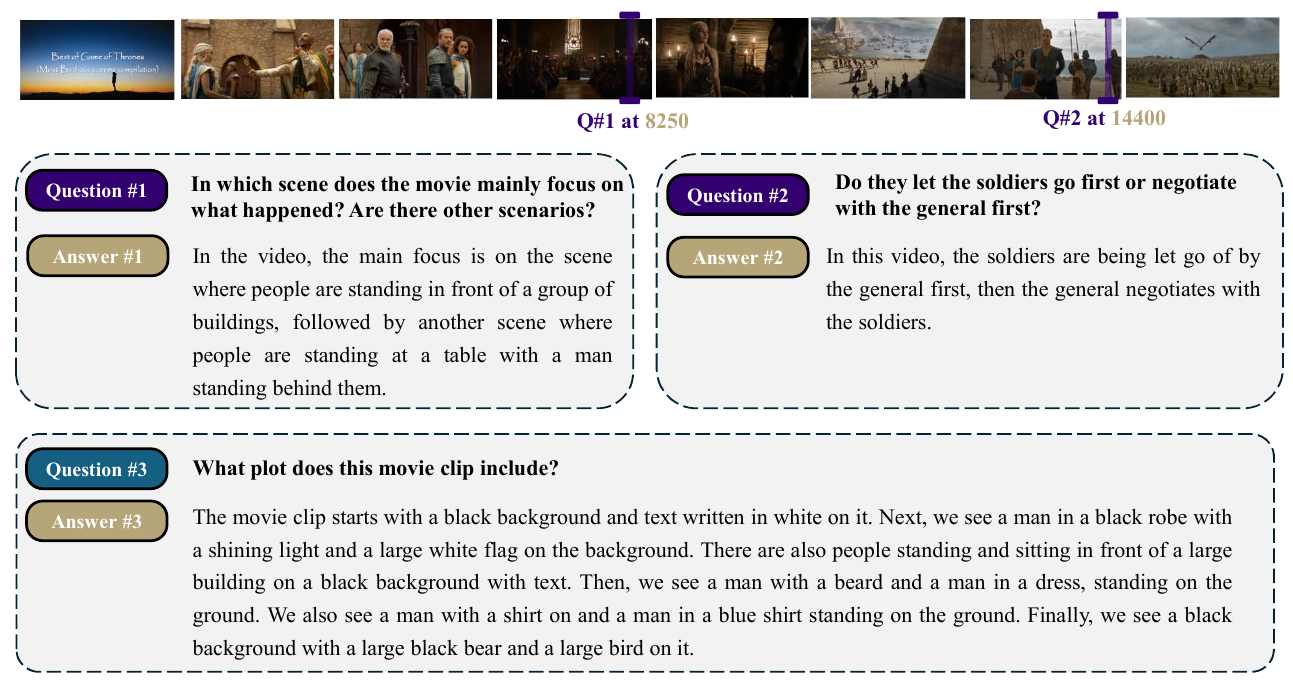}
	\caption{Question and answer about clips from \textit{Game of Thrones}, which tells the epic fantasy tale of power struggles and political intrigue among the Seven Kingdoms, entwined with intricate family relationships, all set against the backdrop of an ancient, mystical threat.}
    \label{fig:case3}
    \vspace{20pt}
    \includegraphics[width=0.95\linewidth]{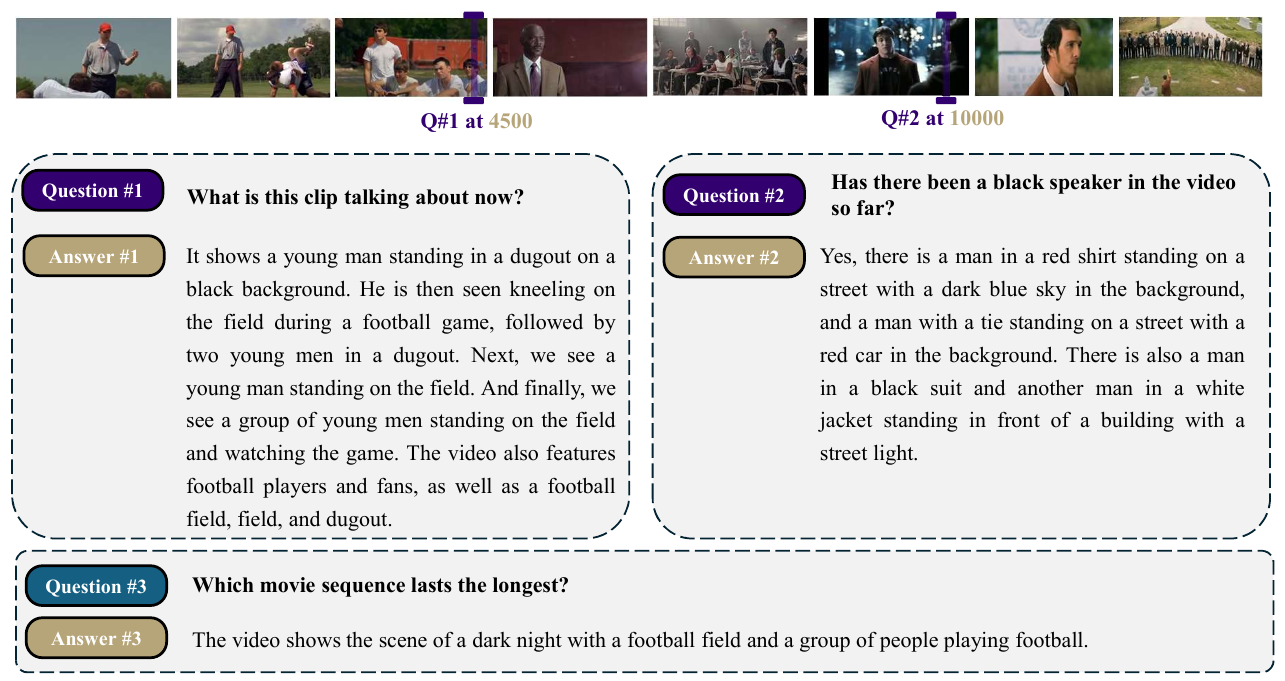}
	\caption{Question and answer about clips from \textit{YouTube}, which contains a compilation of some inspirational movies scenes. This video clip comprises several segments from \textit{The Death Crawl}, \textit{Coach Carter}, \textit{Rocky Balboa}, and \textit{We Are Marshall}, which vary in duration.}
    \label{fig:case4}
 \vspace{20pt}

\end{figure*}